\documentclass[10pt,twocolumn,letterpaper]{article}

\usepackage[pagenumbers]{cvpr} % To force page numbers, e.g. for an arXiv version
\usepackage{tikz}
\usepackage{pgfplots}

\newcommand{\xx}{\mathbf{x}}
\newcommand{\bb}{\mathbf{b}}

\usepackage{amsmath,amsfonts,amssymb,algorithm2e,multirow,mathtools}
\usepackage[accsupp]{axessibility} % Improves PDF readability for those with visual impairments.

\usepackage{stfloats}

\definecolor{cvprblue}{rgb}{0.21,0.49,0.74}
\usepackage[pagebackref,breaklinks,colorlinks,citecolor=cvprblue]{hyperref}

\def\paperID{7380} % *** Enter the Paper ID here
\def\confName{CVPR}
\def\confYear{2024}

\title{Efficient Solution of Point-Line Absolute Pose}

\author{
Petr Hruby\\
ETH Zürich
\and 
Timothy Duff\\
University of Washington
\and 
Marc Pollefeys
\\
ETH Zürich, Microsoft
}

\newcommand{\RR}{\mathbb{R}}

\begin{document}
\maketitle
\begin{abstract}
We revisit certain problems of pose estimation based on 3D--2D correspondences between features which may be points or lines.
Specifically, we address the two previously-studied minimal problems of estimating camera extrinsics from $p \in \{ 1, 2 \}$ point--point correspondences and $l=3-p$ line--line correspondences.
To the best of our knowledge, all of the previously-known practical solutions to these problems required computing the roots of degree $\ge 4$ (univariate) polynomials when $p=2$, or degree $\ge 8$ polynomials when $p=1.$
We describe and implement two elementary solutions which reduce the degrees of the needed polynomials from $4$ to $2$ and from $8$ to $4$, respectively.
We show experimentally that the resulting solvers are numerically stable and fast: when compared to the previous state-of-the art, we may obtain nearly an order of magnitude speedup.
The code is available at \url{https://github.com/petrhruby97/efficient\_absolute}
\end{abstract}    
\section{Introduction}
\label{sec:intro}

\subsection{Motivation}\label{subsec:motivation}

% tim: add citations
The problem of registering images to a known 3D coordinate system plays a crucial role in
applications such as
visual localization~\cite{DBLP:journals/pami/SattlerLK17}, autonomous driving~\cite{DBLP:journals/ivc/HaneHLFFSP17}, and augmented reality~\cite{DBLP:journals/tvcg/VenturaARS14}, as well as in general paradigms like SLAM~\cite{DBLP:journals/access/HuZWYL23} and SfM~\cite{DBLP:conf/cvpr/SchonbergerF16}.
Robust estimators based on RANSAC~\cite{DBLP:journals/cacm/FischlerB81}, or one of its many refinements~\cite{DBLP:journals/pami/RaguramCPMF13,DBLP:journals/pami/BarathM22}, are among the most successful tools for solving these problems.
Such an estimator traditionally relies on a minimal P3P solver~\cite{DBLP:conf/eccv/PerssonN18,DBLP:conf/cvpr/0001YLOA23,DBLP:journals/ijcv/HaralickLON94,DBLP:conf/cvpr/KneipSS11} to efficiently hypothesize poses from putative matches between 3D and 2D points.

The literature on P3P and other purely point-based methods for absolute pose estimation is vast.
Absolute pose estimation from non-point features such as lines~\cite{DBLP:journals/pami/DhomeRLR89,DBLP:conf/icra/RamalingamBS11,DBLP:conf/accv/ZhouYK18,DBLP:journals/pami/Chen91,DBLP:journals/pami/XuZCK17,agostinho2023cvxpnpl}, point-line incidences~\cite{DBLP:journals/pami/FabbriGK21}, and affine correspondences~\cite{DBLP:journals/tvcg/VenturaARS14}, has received comparatively less attention, but remains an active research area.
In particular, solutions relying on both points and lines are of increasing importance, due to the prevalence of both types of feature in man-made environments, as well as several recent advances in the components of 3D reconstruction systems responsible for line detection~\cite{DBLP:conf/cvpr/PautratBLOP23}, matching~\cite{pautrat2023gluestick}, and bundle adjustment~\cite{DBLP:conf/cvpr/LiuYPPL23}.

%\begin{itemize}
%     \color{red}
%     \item Registering images is important in computer vision pipelines i.e. visual localization, 3D reconstruction, and SLAM
%     \item Mention existing point based solvers, line based solvers, and other solvers (Affine correspondence)
%     \item With the new tools (deeplsd, gluestick, limap), there is increasing importance of line based solvers. Mention the previous attempts (Bouaziz, 3Q3)
%     \item Mention that these attempts are suboptimal, since the problems can be decomposed, and, therefore, we do not have to solve as complex problems as they do. Therefore, there is a possibility to obtain faster and more stable solvers.
%     \item Say that this is exactly what we have done, refer to the following sections, and say that we propose new solutions, prove that they have the lowest possible degrees, give an optimized solver that runs faster, and 
    
% \end{itemize}

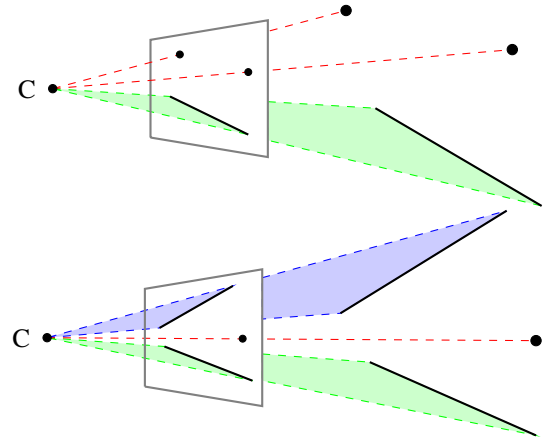
\begin{figure}
    \centering
    \begin{tikzpicture}[scale=1.3]

\filldraw[fill=green!20,draw=green, dashed] (0,0) -- (1.2,-.08) -- (2.0,-.47) -- cycle;
\filldraw[fill=green!20,draw=green, dashed] (2.20,-.133) -- (3.3,-0.2) -- (5,-1.2) -- (2.2,-.528) -- cycle;

\draw[gray, thick] (1,-0.5) -- (2.2,-0.7) -- (2.2,0.7) -- (1,0.5) -- (1,-0.5);
\draw[black, thick] (3.3,-0.2) -- (5,-1.2);

\draw[red, dashed] (0,0) -- (1.3,0.35);
\draw[red, dashed] (2.22,.592) -- (3.0,0.8);
\draw[red, dashed] (0,0) -- (2.0,0.17);
\draw[red, dashed] (2.25,.191) -- (4.7,0.4);
%\draw[green, dashed] (0,0) -- (3.3,-0.2);
%\draw[green, dashed] (0,0) -- (5,-1.2);

\filldraw[black] (3.0,0.8) circle (1.5pt);% node[anchor=east]{C};
\filldraw[black] (4.7,0.4) circle (1.5pt);
\filldraw[black] (0,0) circle (1.2pt) node[anchor=east,xshift=-0.1cm]{C};

\filldraw[black] (1.3,0.35) circle (1.0pt);
\filldraw[black] (2.0,0.17) circle (1.0pt);
\draw[black, thick] (1.2,-0.08) -- (2.0,-0.47);

%\draw[green, thick] (0.2,0.8) -- (0.8,0.2);
%\draw[gray, thick] (-1,-1) -- (2,2);
%\filldraw[black] (0.5,0.5) circle (2pt);% node[anchor=west]{Intersection point};
\end{tikzpicture}
    %\vspace{0.3cm}
    \begin{tikzpicture}[scale=1.3]

\filldraw[fill=green!20,draw=green, dashed] (0,0) -- (1.2,-0.09) -- (2.1,-0.44) -- cycle;
\filldraw[fill=green!20,draw=green, dashed] (2.2,-.1666) -- (3.3,-0.25) -- (5,-1.0) -- (2.2, -.44) -- cycle;
%\filldraw[fill=blue!20,draw=blue, dashed] (0,0) -- (3.0,0.25) -- (4.7,1.3) -- cycle;
\filldraw[fill=blue!20,draw=blue, dashed] (0,0) -- (1.15,.1) -- (1.9,.53) -- cycle;
\filldraw[fill=blue!20,draw=blue, dashed] (2.2, .614) -- (4.7,1.3) --(3.0,0.25) -- (2.2,.191) -- cycle;

\draw[red, dashed] (0,0) -- (2,-0.012);
\draw[red, dashed] (2.25,-.0135) -- (5,-0.03);

\filldraw[black] (0,0) circle (1.2pt) node[anchor=east,xshift=-0.1cm]{C};
\draw[gray, thick] (1,-0.5) -- (2.2,-0.7) -- (2.2,0.7) -- (1,0.5) -- (1,-0.5);
\filldraw[black] (5,-0.03) circle (1.5pt);
\draw[black, thick] (3.3,-0.25) -- (5,-1.0);
\draw[black, thick] (3.0,0.25) -- (4.7,1.3);

\filldraw[black] (2.0,-0.01) circle (1.0pt);
\draw[black, thick] (1.2,-0.09) -- (2.1,-0.44);
\draw[black, thick] (1.15,0.1) -- (1.9,0.53);
\end{tikzpicture}
    \caption{The P2P1L (\textit{Top}) and P1P2L (\textit{Bottom}) problems.}
    \label{fig:teaser}
\end{figure}

\subsection{Contribution}\label{subsec:contribution}

In this paper, we revisit the two minimal problems of absolute pose estimation that combine both points and lines: the Perspective-2-Point-1-Line Problem (or P2P1L,~\Cref{subsec:p2p1l}), and the Perspective-1-Point-2-Line Problem (P1P2L,~\Cref{subsec:p1p2l}.)
See~\Cref{fig:teaser} for illustrations.

In contrast to the purely point or line based minimal problems P3P and P3L, we observe that the existing solutions 
for the ``mixed cases" considered here are suboptimal from both a theoretical and practical point of view.
Consequently, we develop novel solutions to both problems, which optimally exploit their underlying algebraic structure, and exhibit comparable or better performance than the state of the art on simulated and real data---see~\Cref{sec:experiments}.
On the other hand, these solvers are simple to implement, and no knowledge of mathematics beyond elementary algebra is needed to understand them.

To provide further context for our work, we include the following quotation from~\cite{DBLP:conf/icra/RamalingamBS11}: \emph{Although we do
not theoretically prove that our solutions are of the lowest
possible degrees, we believe so because of the following
argument. The best existing solutions for pose estimation
using three points and three lines use 4th and 8th degree
solutions respectively. Since mixed cases are in the middle,
our solutions for (2 points, 1 line) and (1 point, 2 lines) cases use 4th and 8th degree solutions respectively.
Recently, it
was shown using Galois theory that the solutions that use
the lowest possible degrees are the optimal ones [\cite{DBLP:conf/cvpr/NisterHS07}].}

Contrary to the informal reasoning presented above, we claim that the existing solutions to the mixed point--line cases are not optimal.
As far as we know, the only support for this claim appearing in the literature prior to our work occurs in~\cite[\S 3]{galvis1}.
This previous work showed, on the basis of Galois group computation, that the P2P1L and P1P2L problems decompose into simpler subproblems.
However, this theoretical observation was not accompanied by a practical solution method for either problem.
In this paper, we rectify the situation by devising practical solvers for both problems that incorporate these recent insights.

To complete our discussion of what makes a solution \emph{algebraically optimal}, we recall one proposal of such a notion from work of Nist\'{e}r et al.~\cite{DBLP:conf/cvpr/NisterHS07}.
In this work, a restricted class of algorithms $\mathcal{P}_n$ is considered for each natural number $n\ge 1$: an algorithm in $\mathcal{P}_n$ consists of a finite sequence of steps, each of which extracts the roots of some polynomial equation $p(x)=0$, with either $p(x) = x^m -a $ (where $a\in \mathbb{Q}$ and $\deg(p) = m$ can be arbitrary), or with $\deg (p) \le n$ and the coefficients of $p$ belonging to a field containing $\mathbb{Q}$ and any previously-computed roots.
In this setting, a solution is optimal if it leads to solving a polynomial system of the lowest possible degree.
Our proposed solutions immediately establish that P2P1L is $\mathcal{P}_2$-solvable and P1P2L is $\mathcal{P}_4$-solvable.
It should be of little surprise that P2P1L is not $\mathcal{P}_1$-solvable; therefore, we may say that our solution to P2P1L is algebraically optimal.
On the other hand, the problem P1P2L is also $\mathcal{P}_3$-solvable.
This is because \emph{any quartic} equation can be solved by the standard method which reduces the problem to computing the roots of the associated \emph{resolvent cubic}.
Thus, our solution, which computes the roots of a quartic with this same method, is also algebraically optimal in the sense of~\cite{DBLP:conf/cvpr/NisterHS07}.
The same observation, of course, holds for the classical quartic-based methods for solving P3P.
Alternative P3P solvers that directly employ a cubic~\cite{DBLP:conf/eccv/PerssonN18,DBLP:conf/cvpr/0001YLOA23}, despite being superior in practical terms, are not distinguished by the complexity classes $\mathcal{P}_n.$
\Cref{tab:complexity-classes} provides a summary of the algebraic complexities, based on the Galois groups computed in~\cite[\S 3]{galvis1}.

\begin{table}[h]
    \centering
    \begin{tabular}{c|cccc}
problem         & P3P & P2P1L & P1P2L & P3L  \\
\hline
class    & $\mathcal{P}_3$ & $\mathcal{P}_2$ & $\mathcal{P}_3$ & $\mathcal{P}_8$ 
    \end{tabular}
    \caption{For each minimal absolute pose problem with points and line features, the class of polynomial root finding algorithms $\mathcal{P}_n$ that solve the problem with $n$ as small as possible~\cite[\S 2.1]{DBLP:conf/cvpr/NisterHS07}.}
    \label{tab:complexity-classes}
\end{table}

%Sources for intro:
%https://arxiv.org/pdf/2011.08790v5.pdf
%http://sofienbouaziz.com/pdf/Geolocalization_ICRA11.pdf
%https://openaccess.thecvf.com/content/CVPR2023/papers/Ding_Revisiting_the_P3P_Problem_CVPR_2023_paper.pdf

% \begin{itemize}
%     \color{green}
%     \item Teaser: illustrations of the two problems that we are going to solve.
%     \item For each of the solvers separately: add the special configuration from Bouaziz 
%     \item In the section about stabilization: illustrate how the stabilization works.
%     \item There will be plots for time and for stability for every considered solver
%     \item Also, the steps could be recreated into an algorithm
% \end{itemize}

\subsection{Related work}\label{subsec:related-work}

The first solutions to P2P1L and P1P2L were presented by Ramalingam et al.~\cite{DBLP:conf/icra/RamalingamBS11}, reducing the problems to computing the roots of polynomials of degree $4$ and $8$ via careful choices of special reference frames in the world and image.
Although we obtain polynomials of lower degree, these special reference frames are also an ingredient in our approach.

In work subsequent to~\cite{DBLP:conf/icra/RamalingamBS11}, it was observed that both of these problems could be solved using the E3Q3 solver~\cite{DBLP:conf/cvpr/KukelovaHF16}.
This is a highly optimized method for computing the points where three quadric surfaces in $\RR^3$ intersect.
Typically, there are $2^3=8$ such points, by B\'{e}zout's theorem; however, EQ3Q also handles degenerate cases where the number of solutions may drop.
In the follow-up work~\cite{DBLP:conf/accv/ZhouYK18}, a stabilization scheme for E3Q3 was applied to these problems, and experimentally shown to be more accurate than the solvers in~\cite{DBLP:conf/icra/RamalingamBS11}.
Efficient solvers for both problems based on this stabilized E3Q3 are implemented in \texttt{PoseLib}~\cite{PoseLib}.

The idea of using Galois groups to study minimal problems originates from~\cite{DBLP:conf/cvpr/NisterHS07}. The main takeaway from this paper is that a problem with $n$ solutions whose Galois group is the full symmetric group $S_n$ cannot be solved by an algorithm in $\mathcal{P}_{n-1}.$
In the other direction, the main takeaway from~\cite{galvis1} is that if the Galois group is contained in the wreath product $S_{n_1} \wr S_{n_2},$ where $n=n_1 n_2,$ then the problem can be solved with an algorithm in $\mathcal{P}_{\max (n_1, n_2)}.$
A recent work using such an insight to guide the more efficient solution of a \emph{relative pose} estimation problem may be found in~\cite{DBLP:conf/cvpr/HrubyKDOPPL23}.

\section{Minimal solvers}
\label{sec:solvers}

In this section, we introduce our \textit{algebraically optimal} solutions to the P2P1L and P1P2L problems. These problems are depicted in~\Cref{fig:teaser}. As revealed by the Galois group computed in \cite[\S 3]{galvis1}, the problem P2P1L can be reduced to computing the roots of a quadratic equation, and the problem P1P2L can be reduced to a quartic equation. In~\Cref{subsec:p2p1l,subsec:p1p2l}, we turn these insights into explicit solutions to the P2P1L and P12PL problems, respectively.
These solutions work \emph{generically}---they are valid outside of a measure-zero subset of the space of input point-point/line-line correspondences.
In~\Cref{subsec:stabilization}, we describe a method that stabilizes the P1P2L solver of~\Cref{subsec:p1p2l} in a common but non-generic case.
Finally,~\Cref{subsec:coplanar} for a discussion when the point-line configuration is coplanar.

%\begin{itemize}
    %\color{red}
    %\item Do some introducing sentence that says that in this section we provide optimal solutions to problems P2P1L and P1P2L. (and refer to the figures where these problems are shown)
    %\item Mention that there is the monodromy group that can reveal the degree of the problem, as well as the decomposition. (here create an actual itemization or a table)
    %\item For the P2P1L we have 4 solutions, and the problem decomposes into 2*2 => we have to solve a quadratic problem
    %\item For the P1P2L we have 8 solutions, and the problem decomposes into 4*2 => we have to solve a degree 4 problem
    %\item Write what will happen in the following subsections and refer to them at the same time.
%\end{itemize}

\begin{figure}
    \centering
    \begin{tabular}{c c}
       \begin{tikzpicture}[scale=1.5]

\draw[-stealth] (0,0) -- (0,1);
\draw[-stealth] (0,0) -- (1,0);
\draw[-stealth] (0,0) -- (-0.6,-0.6);

\node at (1,-0.15) {$X$};
\node at (-0.15,0.85) {$Y$};
\node at (-0.7,-0.4) {$Z$};

\node at (-0.7,0.6) {\Large $\mathcal{C}_0$};

%\filldraw[fill=green!20,draw=green, dashed] (0,0) -- (3.3,-0.2) -- (5,-1.2) -- cycle;

\fill[blue!20] (1,1) -- (0.15,0.8) -- (0.35,0.35);

\draw[blue, thick, -stealth] (1,1) -- (0.5,1.1) node[yshift=+0.2cm,xshift=0.1cm]{$d_1$};
\draw[blue, thick, -stealth] (1,1) -- (0.9,0.5) node[yshift=+0.2cm,xshift=0.3cm]{$d_2$};
\draw[blue, thick, -stealth] (1,1) -- (0.15,0.8) node[yshift=+0.3cm,xshift=0.1cm]{$d_3$};
\draw[blue, thick, -stealth] (1,1) -- (0.35,0.35) node[yshift=-0.1cm,xshift=0.3cm]{$d_4$};
\filldraw[black] (1,1) circle (1pt) node[anchor=south,xshift=+0.1cm, yshift=0.1cm]{$C^0$};

%\draw[gray, thick] (1,-0.5) -- (2.2,-0.7) -- (2.2,0.7) -- (1,0.5) -- (1,-0.5);
%\filldraw[black] (3.0,0.8) circle (1.5pt);% node[anchor=east]{C};
%\filldraw[black] (4.7,0.4) circle (1.5pt);
%\draw[black, thick] (3.3,-0.2) -- (5,-1.2);

%\draw[red, dashed] (0,0) -- (3.0,0.8);
%\draw[red, dashed] (0,0) -- (4.7,0.4);
%%\draw[green, dashed] (0,0) -- (3.3,-0.2);
%%\draw[green, dashed] (0,0) -- (5,-1.2);

%\filldraw[black] (1.3,0.35) circle (1.0pt);
%\filldraw[black] (2.0,0.17) circle (1.0pt);
%\draw[black, thick] (1.2,-0.08) -- (2.0,-0.47);

%\draw[green, thick] (0.2,0.8) -- (0.8,0.2);
%\draw[gray, thick] (-1,-1) -- (2,2);
%\filldraw[black] (0.5,0.5) circle (2pt);% node[anchor=west]{Intersection point};
\end{tikzpicture}  &  \begin{tikzpicture}[scale=1.5]

\draw[-stealth] (0,0) -- (0,1);
\draw[-stealth] (0,0) -- (1,0);
\draw[-stealth] (0,0) -- (-0.6,-0.6);

\node at (1,-0.15) {$X$};
\node at (-0.15,0.85) {$Y$};
\node at (-0.7,-0.4) {$Z$};

\node at (-0.6,0.6) {\Large $\mathcal{W}_0$};

\filldraw[red] (0.3,0.85) circle (1pt) node[anchor=north,xshift=-0.1cm, yshift=0.0cm]{$P_1^0$};
\filldraw[red] (0.9,0.5) circle (1pt) node[anchor=north,xshift=-0.1cm, yshift=0.0cm]{$P_2^0$};

\draw[black, thick] (0.4,1.1) -- (1.1,0.64);
\filldraw[red] (0.55,1.0) circle (1pt) node[anchor=south,xshift=-0.0cm, yshift=0.0cm]{$L_3^0$};
\filldraw[red] (0.85,0.8) circle (1pt) node[anchor=west,xshift=-0.1cm, yshift=0.3cm]{$L_4^0$};

%\draw[blue, thick, -stealth] (1,1) -- (0.9,0.5) node[yshift=+0.2cm,xshift=0.3cm]{$d_2$};
%\draw[blue, thick, -stealth] (1,1) -- (0.15,0.8) node[yshift=+0.3cm,xshift=0.1cm]{$d_3$};
%\draw[blue, thick, -stealth] (1,1) -- (0.35,0.35) node[yshift=-0.1cm,xshift=0.3cm]{$d_4$};

\end{tikzpicture}  \\
       \begin{tikzpicture}[scale=1.5]

\fill[blue!20] (0.6,0.6) -- (-0.1,-0.1) -- (0.45,-0.2);

\draw[-stealth] (0,0) -- (0,1);
\draw[-stealth] (0,0) -- (1,0);
\draw[-stealth] (0,0) -- (-0.6,-0.6);

\node at (1,-0.15) {$X$};
\node at (-0.15,0.85) {$Y$};
\node at (-0.7,-0.4) {$Z$};

\node at (-0.7,0.6) {\Large $\mathcal{C}_1$};

\draw[blue, thick, -stealth] (0.6,0.6) -- (0.1,0.5) node[yshift=+0.3cm,xshift=0.2cm]{$D_1$};
\draw[blue, thick, -stealth] (0.6,0.6) -- (0.78,0.2) node[yshift=+0.2cm,xshift=0.3cm]{$D_2$};
\draw[blue, thick, -stealth] (0.6,0.6) -- (-0.1,-0.1) node[yshift=+0.3cm,xshift=-0.15cm]{$D_3$};
\draw[blue, thick, -stealth] (0.6,0.6) -- (0.45,-0.2) node[yshift=-0.1cm,xshift=0.3cm]{$D_4$};
\filldraw[black] (0.6,0.6) circle (1pt) node[anchor=south,xshift=+0.1cm, yshift=0.1cm]{$C$};

\filldraw[black] (0,0) circle (1pt);
\filldraw[black] (0.49,0) circle (1pt);
\filldraw[black] (0.25,0.52) circle
(1pt);
\filldraw[black] (0.71,0.35) circle
(1pt);

\end{tikzpicture}  &  \begin{tikzpicture}[scale=1.5]

\draw[-stealth] (0,0) -- (0,1);
\draw[-stealth] (0,0) -- (1,0);
\draw[-stealth] (0,0) -- (-0.6,-0.6);

\node at (1,-0.15) {$X$};
\node at (-0.15,0.85) {$Y$};
\node at (-0.7,-0.4) {$Z$};
\node at (-0.6,0.6) {\Large $\mathcal{W}_1$};

\filldraw[red] (0.0,0.0) circle (1pt) node[anchor=north,xshift=+0.1cm, yshift=0.0cm]{$P_1$};
\filldraw[red] (0.8,0.0) circle (1pt) node[anchor=north,xshift=-0.1cm, yshift=0.0cm]{$P_2$};

\draw[black, thick] (0.1,0.38) -- (0.9,0.24);
\filldraw[red] (0.18,0.36) circle (1pt) node[anchor=south,xshift=-0.0cm, yshift=0.0cm]{$L_3$};
\filldraw[red] (0.7,0.27) circle (1pt) node[anchor=west,xshift=-0.1cm, yshift=0.3cm]{$L_4$};

\end{tikzpicture}
    \end{tabular}
    \vspace{-0.2cm}
    \caption{\textbf{Special reference frame for P2P1L.} Camera frame $\mathcal{C}_0$ is transformed to $\mathcal{C}_1$, world frame $\mathcal{W}_0$ to $\mathcal{W}_1$. See text for details.}
    \label{fig:p2p1l_frame}
\end{figure}
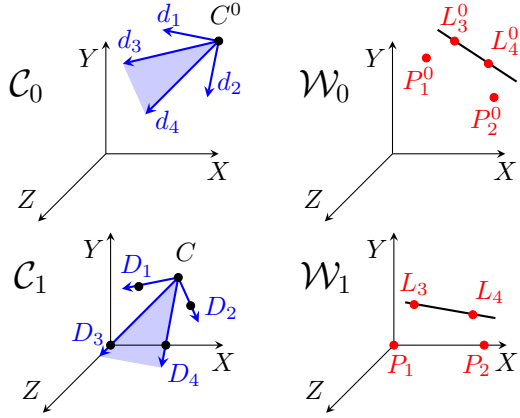

\subsection{P2P1L}\label{subsec:p2p1l}

%\begin{itemize}
%    \color{red}
%    \item Formalize the \textbf{Problem Statement}: Consider two 2D points $P_i$ ...
%    \item When talking about the special frame, refer to the image which shows it. And specifically state that we transform the points to it. (and add an arrow to the image)
%    %\item Make the formulae fit the page
%    \item ? change the summary into an algorithm / list
%    \item add some sentence at the end, say what happens, refer to the experiments or something like that
%\end{itemize}

Here, we provide the algebraically optimal solution to the P2P1L problem. We mostly follow the notation of~\cite{DBLP:conf/icra/RamalingamBS11}.%, in which special reference frames are chosen to simplify the equations.

\textbf{Problem Statement}: Consider $P_1^0, P_2^0 \in \RR^3$, two 3D points, and their 2D projections under an unknown calibrated camera.
%; denote by $d_1, d_2 \in \RR^3$ these projections in homogeneous coordinates. 
Consider also a 3D line spanned by points $L_3^0, L_4^0 \in \RR^3$ and its projection. %spanned by points $d_3, d_4\in \RR^3$.
The goal is to recover the unknown camera matrix, whose center we denote by $C^0$.

Our first step is to rigidly transform the input into the special reference frame introduced in~\cite{DBLP:conf/icra/RamalingamBS11}. Transforming to this special frame simplifies the equations and helps to reveal the algebraically optimal solution.
%The special frame is illustrated in the bottom row of Figure~\ref{fig:p2p1l_frame}.
%to simplify the equations. This  
\Cref{fig:p2p1l_frame} illustrates the input reference frame $\mathcal{C}_0$ / $\mathcal{W}_0$ (top), and the special frame $\mathcal{C}_1$ / $\mathcal{W}_1$ (bottom), which are related as follows:
\begin{itemize}
    \item In the world frame $\mathcal{W}_1$, the 3D points take the form
    \begin{equation}\label{eq:world-pts-transformed}
P_1 = \begin{pmatrix}
    0 & 0 & 0
\end{pmatrix}^T,
\quad 
P_2 = 
\begin{pmatrix}
    X_2 & 0 & 0
\end{pmatrix}^T,
    \end{equation}
and the 3D line is spanned by two points of the form
    \begin{equation}\label{eq:world-ln-transformed}
L_3 = \begin{pmatrix}
    X_3 & Y_3 & 0
\end{pmatrix}^T,
\quad 
L_4 = 
\begin{pmatrix}
    X_4 & Y_4 & Z_4
\end{pmatrix}^T.
    \end{equation}
    \item To find the transformation $\mathcal{W}_0 \rightarrow \mathcal{W}_1$, we set 
$$P_1 = \begin{pmatrix} 0 & 0 & 0 \end{pmatrix}^T,\quad  P_2 = \begin{pmatrix} \lVert P_1^0 - P_2^2 \rVert & 0 & 0 \end{pmatrix}^T,$$ 
$$L_3 = \begin{pmatrix} X_3 & Y_3 & 0 \end{pmatrix}^T, \text{ w/ }\, X_3 = (L_3^0-P_1)^T \frac{P_2-P_1}{\lVert P_2-P_1 \rVert},$$
$$Y_3 = \Big\lVert L_3^0 - P_1 - X_3\cdot\frac{P_2-P_1}{\lVert P_2-P_1 \rVert} \Big\rVert .$$ 
%and $L_3 = \begin{pmatrix} X_3 & Y_3 & 0 \end{pmatrix}^T$.
We transform the rays $\overrightarrow{P_1^0 P_2^0}$ and $\overrightarrow{P_1^0 L_3^0}$ to $\overrightarrow{P_1 P_2}$ and $\overrightarrow{P_1 L_3},$ respectively, via suitable translation and rotation.
    \item In the camera frame, the camera center is fixed at 
    \begin{equation}\label{eq:C}
C = \begin{pmatrix}
    0 & 0 & -1 
\end{pmatrix}^T,
    \end{equation}
    the 2D points are projections of points of the form
    \begin{equation}\label{eq:cam-pts-transformed}
D_1 = \begin{pmatrix}
    a_1 & b_1 & 0
\end{pmatrix}^T,
\quad 
D_2 = 
\begin{pmatrix}
    a_2 & b_2 & 0
\end{pmatrix}^T,
    \end{equation}
and the 2D line is the projection of $\overline{D_3 D_4},$ where
    \begin{equation}\label{eq:cam-ln-transformed}
D_3 = \begin{pmatrix}
    0 & 0 & 0
\end{pmatrix}^T,
\quad 
D_4 = 
\begin{pmatrix}
    1 & 0 & 0
\end{pmatrix}^T.
\end{equation}
\item To find the transformation $\mathcal{C}_0 \rightarrow \mathcal{C}_1$, let $D_3^0$ and $D_4^0$ be the homogeneous coordinates of two distinct points along the given line.
Independently of $C^0,$ we may rigidly transform the rays $d_3 := \overrightarrow{C^0 D_3^0}$, and $d_4 := \overrightarrow{C^0 D_3^0}$ to $\overrightarrow{C D_3}$ and $\overrightarrow{C D_4},$ by suitable rotation and translation.
\end{itemize}
Let us now write $R= (R_{i j})_{1 \le i, j \le 3}$, $T = (T_i)_{1\le i \le 3}$ for the unknown camera pose in this special reference frame.
Our projection constraints can then be written as
\begin{align}\label{eq:pt-pt-equation}
    (D_i-C) \times (R P_i + T)=0, \quad i \in \{1, 2\},\\
    \label{eq:ln-ln-equation}
    (D_3 \times D_4)^T (R L_i + T) = 0, \quad i \in \{3, 4\}.\end{align}
This gives a system of equations in $12$ unknowns.
However, in what follows, we consider a system of equations in a smaller set of unknowns, namely
\begin{equation}\label{eq:unknown-vector-p2p1l}
\xx^T = \begin{pmatrix}
R_{1 1}&
R_{2 1} &
R_{3 1} &
R_{2 2} &
R_{2 3} &
T_1&
T_2&
T_3
\end{pmatrix}.
\end{equation}
The entries of $\xx$ are constrained by
\begin{align}\label{eq:ortho}
R_{1 1}^2 + R_{2 1}^2 + R_{3 1}^2 = 
R_{2 1}^2 + R_{2 2}^2 + R_{2 3}^2 &= 1.
\end{align}
Thus, the entries of $\xx $ specify a translation vector and a partially-filled rotation matrix, which can be uniquely completed to a rotation matrix using the formulae
\begin{align}\label{eq:recover-remaining-R}
\begin{pmatrix}
R_{1 2}\\
R_{1 3} \\
R_{3 2} \\
R_{3 3} 
\end{pmatrix} =
(R_{2 2}^2 + R_{2 3}^2)^{-1}
\cdot 
\begin{pmatrix}
-R_{1 1} R_{2 1} R_{2 2} + R_{2 3} R_{3 1}\\
-R_{1 1} R_{2 1} R_{2 3} - R_{2 2} R_{3 1} \\
-R_{2 1} R_{2 2} R_{3 1} - R_{1 1} R_{2 3} \\
-R_{2 1} R_{2 3} R_{3 1} + R_{1 1} R_{2 2}
\end{pmatrix},
\end{align}
subject to the genericity condition $R_{22}^2 + R_{23}^2 \ne 0$, ie.~when $R$ is not a rotation in the $xz$-plane.\footnote{If $R$ is known to be a plane rotation, two generic point-point correspondences suffice to recover $R$ and $T.$}
Hence, we focus on recovering the entries of $\xx $. 
From 2 out of the 3 redundant constraints from \eqref{eq:pt-pt-equation} together with \eqref{eq:ln-ln-equation}, we obtain linear constraints on $\xx$ (\cite[eq.~(4)--(5)]{DBLP:conf/icra/RamalingamBS11}),
\begin{align}\label{eq:linear_constraints-p2p1l}
A \xx &= \bb , \quad \text{where} \nonumber \\
A &=  \begin{psmallmatrix}
0 & 0 & 0 & 0 & 0 & - b_1 & a_1 & 0\\
0 & 0 & 0 & 0 & 0 & 0 & -1 & b_1\\
-b_2 X_2 & a_2 X_2 & 0 & 0 & 0 & - b_2 & a_2 & 0\\
0 & -X_2 & b_2 X_2 & 0 & 0 & 0 & -1 & b_2\\
0 & X_3 & 0 & Y_3 & 0 & 0 & 1 & 0\\
0 & X_4 & 0 & Y_4 & Z_4 & 0 & 1 & 0
\end{psmallmatrix},  \nonumber  \\
\bb^T &= \begin{pmatrix}
0&
-b_1 &
0 &
-b_2 &
0 &
0
\end{pmatrix}. 
\end{align}
%As previously observed in \emph{loc.~cit.}, the linear equations~\eqref{eq:linear_constraints-p2p1l} follow directly from the special choice of reference frame and the basic equations~\eqref{eq:pt-pt-equation}--~\eqref{eq:ln-ln-equation}.
We may use these linear equations to solve for the translation $T$ in terms of $R$ and the problem data as follows:
\begin{align}\label{eq:T-backsolve}
T_1 &= a_1 B,
\quad T_2 = b_1 B,
\quad T_3 = -1 + B,
\quad \text{where} \nonumber \\
B &= 
\displaystyle\frac{X_2 ( a_2 R_{2 1} - b_2 R_{1 1})}{b_2 a_1 - b_1 a_2}.
\end{align}
Moreover, from~\eqref{eq:linear_constraints-p2p1l} we may use $R_{1 1}$ and $R_{2 1}$, to express the remaining entries of $R$ as 
\begin{align}\label{eq:R-backsolve}
R_{3 1} &= \displaystyle\frac{(a_1 -a_2) R_{2 1} + (b_2-b_1) R_{1 1})}{b_2 a_1 - b_1 a_2}, \nonumber \\
R_{2 2} &= 
\displaystyle\frac{b_1 b_2 X_2 R_{1 1} + (b_1 a_2 X_2 - b_1 a_2 X_2 - b_2 a_1 X_3) R_{2 1}}{Y_3 (b_2 a_1 - b_1 a_2)}
, \nonumber \\[.01\textwidth]
R_{2 3} &= 
\displaystyle\frac{b_1b_2X_2 (Y_3 - Y_4) R_{1 1}
}{Y_3 Z_4 (b_2 a_1 - b_1 a_2)} \nonumber \\[.01\textwidth]
&+ 
\left(\displaystyle\frac{X_3 Y_4 - X_4 Y_3}{Y_3 Z_4}
+ \displaystyle\frac{b_2 a_1 X_3 Y_4 - b_1 a_2 X_2 Y_3}{Y_3 Z_4 (b_2 a_1 - b_1 a_2)} \right) R_{2 1}.
\end{align}
Substituting~\eqref{eq:R-backsolve} into~\eqref{eq:ortho},
we obtain two bivariate quadratic constraints in $R_{1 1}$ and $R_{2 1}$.
In matrix form,
\begin{equation}\label{eq:two-quadratics}
\begin{pmatrix}
c_1 & c_2 & c_3 \\
d_1 & d_2 & d_3
\end{pmatrix}
\cdot 
\begin{pmatrix}
R_{1 1}^2 \\
R_{1 1} R_{2 1} \\
R_{2 1}^2 
\end{pmatrix}
= 
\begin{pmatrix}
1 \\
1
\end{pmatrix}
,
\end{equation}
where the coefficients $c_{1},c_2,c_3,d_1,d_2,d_3$ are rational functions of the problem data.
Applying the change of variables 
\begin{equation}\label{eq:cov}
u = R_{1 1}^2, \quad v = R_{2 1} / R_{1 1},
\end{equation}
and subtracting the two equations in~\eqref{eq:two-quadratics}, we obtain
\[
(c_1 - d_1) u + (c_2-d_2) uv + (c_3 - d_3) u v^2 = 0.
\]
Assuming $u\ne 0,$ we therefore have the univariate quadratic equation in $v$
\begin{equation}\label{eq:v-quadratic}
(c_1 - d_1) + (c_2-d_2) v + (c_3 - d_3) v^2 = 0.
\end{equation}
If $v$ is one of the roots of~\eqref{eq:v-quadratic}, we may recover a corresponding value for $u$ using one of the equations in~\eqref{eq:two-quadratics}, eg.
\begin{equation}\label{eq:u-linear}
u = (c_1 + c_2 v + c_3 v^2)^{-1},
\quad 
\text{or}
\quad 
u = (d_1 + d_2 v + d_3 v^2)^{-1}.
\end{equation}

To summarize, we provide the outline of steps for solving the P2P1L absolute pose problem in Figure~\ref{alg:p2p1l}.

\begin{figure}[t]
\begin{enumerate}
%\hrule 
    \item[(1)] Transform data in world and camera frames to the special reference frames satisfying~\eqref{eq:world-pts-transformed},~\eqref{eq:world-ln-transformed},~\eqref{eq:C},~\eqref{eq:cam-pts-transformed},~\eqref{eq:cam-ln-transformed}.
    \item[(2)] Compute up to 2 solutions in $v$ to the quadratic~\eqref{eq:v-quadratic}.
    \item[(3)] For each root $v$ in step (2), recover a corresponding value of $u$ using either of the linear expressions in~\eqref{eq:u-linear}.
    \item[(4)] Recover up to $4$ solutions in $\{ R_{1 1}, R_{2 1} \}$ using~\eqref{eq:cov}.
    \item[(5)] Recover $R$ and $T$ using~\eqref{eq:R-backsolve},~\eqref{eq:T-backsolve},~\eqref{eq:recover-remaining-R}.
    \item[(6)] Reverse the transformation applied in step (1).
    \hrule 
\end{enumerate}
\caption{Generic \textbf{P2P1L solver.} See Sec.~\ref{subsec:p2p1l} for details.}\label{alg:p2p1l}
\end{figure}
Algebraically, the nontrivial steps in Figure~\ref{alg:p2p1l} are the second and fourth, which require solving a univariate quadratic equation and computing square roots, respectively.

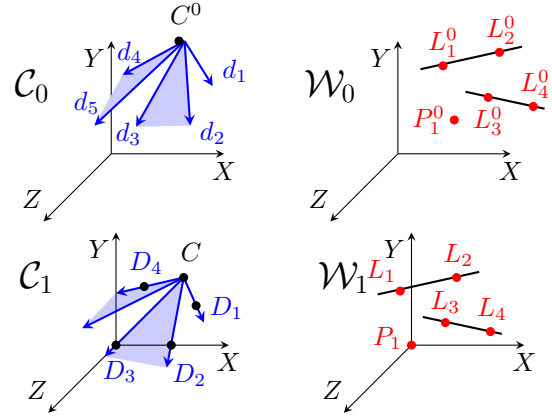
\begin{figure}
    \centering
    \begin{tabular}{c c}
       \begin{tikzpicture}[scale=1.5]

\draw[-stealth] (0,0) -- (0,1);
\draw[-stealth] (0,0) -- (1,0);
\draw[-stealth] (0,0) -- (-0.6,-0.6);

\node at (1,-0.15) {$X$};
\node at (-0.15,0.85) {$Y$};
\node at (-0.7,-0.4) {$Z$};

\node at (-0.7,0.6) {\Large $\mathcal{C}_0$};

\fill[blue!20] (0.65,1) -- (0.7,0.25) -- (0.22,0.24);
\fill[blue!20] (0.65,1) -- (0.1,0.7) -- (-0.15,0.25);

\draw[blue, thick, -stealth] (0.65,1) -- (0.9,0.6) node[yshift=+0.2cm,xshift=0.3cm]{$d_1$};

\draw[blue, thick, -stealth] (0.65,1) -- (0.7,0.25) node[yshift=-0.1cm,xshift=0.3cm]{$d_2$};
\draw[blue, thick, -stealth] (0.65,1) -- (0.22,0.24) node[yshift=-0.1cm,xshift=-0.1cm]{$d_3$};

\draw[blue, thick, -stealth] (0.65,1) -- (0.1,0.7) node[yshift=+0.3cm,xshift=0.1cm]{$d_4$};
\draw[blue, thick, -stealth] (0.65,1) -- (-0.15,0.25) node[yshift=0.3cm,xshift=-0.1cm]{$d_5$};

\filldraw[black] (0.6,1) circle (1pt) node[anchor=south,xshift=+0.1cm, yshift=0.1cm]{$C^0$};

\end{tikzpicture}  &  \begin{tikzpicture}[scale=1.5]

\draw[-stealth] (0,0) -- (0,1);
\draw[-stealth] (0,0) -- (1,0);
\draw[-stealth] (0,0) -- (-0.6,-0.6);

\node at (1,-0.15) {$X$};
\node at (-0.15,0.85) {$Y$};
\node at (-0.7,-0.4) {$Z$};

\node at (-0.6,0.6) {\Large $\mathcal{W}_0$};

\filldraw[red] (0.5,0.3) circle (1pt) node[anchor=east,xshift=-0.0cm, yshift=0.0cm]{$P_1^0$};

\draw[black, thick] (0.2,0.75) -- (1.1,0.95);
\filldraw[red] (0.4,0.78) circle (1pt) node[anchor=south,xshift=-0.0cm, yshift=0.0cm]{$L_1^0$};
\filldraw[red] (0.9,0.9) circle (1pt) node[anchor=west,xshift=-0.3cm, yshift=0.3cm]{$L_2^0$};

\draw[black, thick] (0.6,0.55) -- (1.3,0.4);
\filldraw[red] (0.8,0.5) circle (1pt) node[anchor=north,xshift=-0.0cm, yshift=0.0cm]{$L_3^0$};
\filldraw[red] (1.2,0.42) circle (1pt) node[anchor=west,xshift=-0.3cm, yshift=0.3cm]{$L_4^0$};

\end{tikzpicture}  \\
       \begin{tikzpicture}[scale=1.5]

\fill[blue!20] (0.6,0.6) -- (-0.1,-0.1) -- (0.45,-0.2);
\fill[blue!20] (0.6,0.6) -- (0.0,0.46) -- (-0.3,0.15);

\draw[-stealth] (0,0) -- (0,1);
\draw[-stealth] (0,0) -- (1,0);
\draw[-stealth] (0,0) -- (-0.6,-0.6);

\node at (1,-0.15) {$X$};
\node at (-0.15,0.85) {$Y$};
\node at (-0.7,-0.4) {$Z$};

\node at (-0.7,0.6) {\Large $\mathcal{C}_1$};

\draw[blue, thick, -stealth] (0.6,0.6) -- (0.0,0.46) node[yshift=+0.4cm,xshift=0.4cm]{$D_4$};
\draw[blue, thick, -stealth] (0.6,0.6) -- (-0.3,0.15);% node[yshift=+0.4cm,xshift=0.4cm]{$D_4$};
\draw[blue, thick, -stealth] (0.6,0.6) -- (0.78,0.2) node[yshift=+0.2cm,xshift=0.3cm]{$D_1$};
\draw[blue, thick, -stealth] (0.6,0.6) -- (-0.1,-0.1) node[yshift=-0.2cm,xshift=0.2cm]{$D_3$};
\draw[blue, thick, -stealth] (0.6,0.6) -- (0.45,-0.2) node[yshift=-0.1cm,xshift=0.3cm]{$D_2$};
\filldraw[black] (0.6,0.6) circle (1pt) node[anchor=south,xshift=+0.1cm, yshift=0.1cm]{$C$};

\filldraw[black] (0,0) circle (1pt);
\filldraw[black] (0.49,0) circle (1pt);
\filldraw[black] (0.25,0.52) circle
(1pt);
\filldraw[black] (0.71,0.35) circle
(1pt);

\end{tikzpicture}  &  \begin{tikzpicture}[scale=1.5]

\draw[-stealth] (0,0) -- (0,1);
\draw[-stealth] (0,0) -- (1,0);
\draw[-stealth] (0,0) -- (-0.6,-0.6);

\node at (1,-0.15) {$X$};
\node at (-0.15,0.85) {$Y$};
\node at (-0.7,-0.4) {$Z$};

\node at (-0.6,0.6) {\Large $\mathcal{W}_1$};

\filldraw[red] (0.0,0.0) circle (1pt) node[anchor=east,xshift=-0.0cm, yshift=0.1cm]{$P_1$};

\draw[black, thick] (-0.3,0.45) -- (0.6,0.65);
\filldraw[red] (-0.1,0.48) circle (1pt) node[anchor=south,xshift=-0.25cm, yshift=0.0cm]{$L_1$};
\filldraw[red] (0.4,0.6) circle (1pt) node[anchor=west,xshift=-0.3cm, yshift=0.3cm]{$L_2$};

\draw[black, thick] (0.1,0.25) -- (0.8,0.1);
\filldraw[red] (0.3,0.2) circle (1pt) node[anchor=south,xshift=-0.0cm, yshift=0.0cm]{$L_3$};
\filldraw[red] (0.7,0.12) circle (1pt) node[anchor=west,xshift=-0.3cm, yshift=0.3cm]{$L_4$};

\end{tikzpicture}
    \end{tabular}
    \vspace{-0.2cm}
    \caption{\textbf{Special reference frame for P1P2L.} Camera frame $\mathcal{C}_0$ is transformed to $\mathcal{C}_1$, world frame $\mathcal{W}_0$ to $\mathcal{W}_1$. See text for details.}
    \label{fig:p1p2l_frame}
\end{figure}

\subsection{P1P2L}\label{subsec:p1p2l}
%\begin{itemize}
%    \color{red}
%    \item Formalize the \textbf{Problem Statement}: Consider one 2D point $P_i$ ...
%    \item When talking about the special frame, refer to the image which shows it. And specifically state that we transform %the points to it. (and add an arrow to the image)
%    \item Make the formulae fit the page
%    \item add a summary in the form of an algorithm / list
%    \item add some 
%\end{itemize}

Here, we provide the algebraically optimal solution to the P1P2L problem. We mostly follow the notation of~\cite{DBLP:conf/icra/RamalingamBS11}.%, in which special reference frames are chosen to simplify the equations.

\textbf{Problem Statement}: Let us consider a 3D point $P_1^0 \in \RR^3$, and its homogeneous 2D projection $D_1 \in \RR^3$ under an unknown calibrated camera matrix.
Let us also consider two 3D lines---the first spanned by points $L_1, L_2 \in \RR^3$, and the second one is spanned by $L_3, L_4 \in \RR^3$---and both of their corresponding projections. 
Our task is to recover the camera matrix from the given point-point correspondence and line-line correspondences. 

Much like the P2P1L solver, our P1P2L solver begins by transforming the input data into a special reference frame, as illustrated in~\Cref{fig:p1p2l_frame}.
Specifically,
\begin{itemize}
    \item In the world frame, the 3D point takes the form
    \begin{equation}\label{eq:world-pt-transformed}
        P_1 = \begin{pmatrix}
            0 & 0 & 0
        \end{pmatrix}^T.
    \end{equation}
    The first line is spanned by points 
    \begin{equation}\label{eq:world-ln1-transformed}
        L_1 = \begin{pmatrix}
            X_1 & Y_1 & Z_1
        \end{pmatrix}^T
        , \quad
        L_2 = \begin{pmatrix}
            X_2 & Y_2 & Z_2
        \end{pmatrix}^T,
    \end{equation}
    and the second line is spanned by points
    \begin{equation}\label{eq:world-ln2-transformed}
        L_3 = \begin{pmatrix}
            X_3 & Y_3 & Z_3
        \end{pmatrix}^T
        , \quad L_4 = \begin{pmatrix}
            X_4 & Y_4 & Z_4
        \end{pmatrix}^T.
    \end{equation}
    \item For $\mathcal{W}_0 \rightarrow \mathcal{W}_1$, we simply translate $P_1^0$ to $P_1$.
    \item The world frame may still be rotated freely. We may use the strategy of~\Cref{subsec:stabilization}, which makes the solver stable for a larger class of non-coplanar scenes.
    However, if $P_1, \overline{L_1 L_2}, $ and $\overline{L_3 L_4}$ are coplanar, no rotation is recommended.
   \item In the camera frame, the camera center is fixed at
    \begin{equation}\label{eq:C2}
        C=\begin{pmatrix}
            0 & 0 & -1
        \end{pmatrix}^T.    
    \end{equation}    
    The image point is the projection of a point of the form 
    \begin{equation}\label{eq:cam-pt-transformed}
        D_1 = \begin{pmatrix}
            a_1 & b_1 & 0
        \end{pmatrix}^T,
    \end{equation}
    the first line in the image is the projection of a line spanned by points of the form
    \begin{equation}\label{eq:cam-ln1-transformed}
        D_2 = \begin{pmatrix}
            a_2 & 0 & 0
        \end{pmatrix}^T,
        \quad 
        D_3 = \begin{pmatrix}
            0 & 0 & 0
        \end{pmatrix}^T,
    \end{equation}
    and the second line in the image is the projection of a line spanned by points of the form
    \begin{equation}\label{eq:cam-ln2-transformed}
        D_4 = \begin{pmatrix}
            a_4 & b_4 & 0
        \end{pmatrix}^T,
        \quad 
        D_5 = \begin{pmatrix}
            a_5 & b_5 & 0
        \end{pmatrix}^T.
    \end{equation}

    \item The homogeneous coordinates of the lines in the camera frame are $n_1 = d_2 \times d_3$, $n_2 = d_4 \times d_5$, where the rays $d_2 \ne d_3, d_4 \ne d_5$ to meet the lines in distinct points.
    \item To find the transformation $\mathcal{C}_0 \rightarrow \mathcal{C}_1$, we define 
    $$d_{12} = n_1 \times n_2, \quad D_2^0 = C^0 + \frac{d_2}{d_2^T d_{12}}, \quad D_3^0 = C^0 + d_3, $$
    $$D_2 = \begin{pmatrix} \tan(\cos^{-1}(d_2^T d_{12})) & 0 & 0 \end{pmatrix}^T, \quad D_3 = \begin{pmatrix} 0 & 0 & 0 \end{pmatrix}^T,$$ and, independently of $C^0$, rigidly transform the rays $d_2 := \overrightarrow{C^0 D_2^0}$ and $d_3 := \overrightarrow{C^0 D_3^0}$ to $\overrightarrow{C^0 D_2^0}$ and $\overrightarrow{C D_3}.$
    %To find the transformation $\mathcal{C}_0 \rightarrow \mathcal{C}_1$, we define $D_3^0 = C^0 + d_3$, $D_4^0 = C^0 + \frac{d_4}{d_3 \cdot d_4}$ and set $D_3 = \begin{pmatrix} 0 & 0 & 0 \end{pmatrix}^T$, $D_4 = \begin{pmatrix} \tan(\cos^{-1}(d_3 \cdot d_4)) & 0 & 0 \end{pmatrix}^T$. Transformation $\mathcal{C}_0 \rightarrow \mathcal{C}_1$ maps $(C^0, D_3^0, D_4^0)$ to $(C, D_3, D_4)$.
\end{itemize}

In the special reference frame, we now write the pose as $(R_{ij})_{1 \leq i,j \leq 3}$, $(T_{i})_{1 \leq i \leq 3}$. 
The projection constraints can be formulated as:
\begin{equation}\label{eq:pt-pt-equation-P1P2L}
    (D_1-C) \times (R P_1 + T)=0,
    %D_1 \sim R P_1 + T,
\end{equation}
\begin{equation}\label{eq:ln-ln-equation-01}
    (D_2 \times D_3)^T (R L_i + T) = 0, i \in \{1, 2\},
\end{equation}
\begin{equation}\label{eq:ln-ln-equation-02}
    (D_4 \times D_5)^T (R L_i + T) = 0, i \in \{3, 4\}.
\end{equation}
Analagously to~\eqref{eq:linear_constraints-p2p1l}, 
we have a system of linear equations $A \xx = \bb,$ obtained by picking 2 out of the 3 redundant constraints from \eqref{eq:pt-pt-equation-P1P2L} together with \eqref{eq:ln-ln-equation-01}, \eqref{eq:ln-ln-equation-02} , where now (cf.~\cite[eq.~7--8]{DBLP:conf/icra/RamalingamBS11})
\begin{align*}
% A^T &= \begin{pmatrix}
%         0 & 0 & 0 & 0 & -b_4 X_3 & -b_4 X_4\\
%         0 & 0 & 0 & 0 & -b_4 Y_3 & -b_4 Z_4\\
%         0 & 0 & X_1 & X_2 & a_4 X_3 & a_4 X_4\\
%         0 & 0 & Y_1 & Y_2 & a_4 Y_3 & a_4 Y_4\\
%         0 & 0 & Y_1 & Y_2 & a_4 Y_3 & a_4 Y_4\\
%         0 & 0 & Z_1 & Z_2 & a_4 Z_3 & a_4 Z_4\\
% -b_1 & 0 & 0 & 0 & -b_4 & - b_4 \\
% a_1 & -1 & 1 & 1 & a_4 & a_4 \\
% 0 & b_1 & 0 & 0 & 0 & 0 
%     \end{pmatrix},\\
A &= \begin{psmallmatrix}
0 & 0 & 0 & 0 & 0 & 0 & -b_1 & a_1 & 0\\
0 & 0 & 0 & 0 & 0 & 0 & 0 & -1 & b_1\\
0 & 0 & X_1 & Y_1 & Y_1 & Z_1 & 0 & 1 & 0\\
0 & 0 & X_2 & Y_2 & Y_2 & Z_2 & 0 & 1 & 0\\
-b_4 X_3 & -b_4 Y_3 & a_4 X_3 & a_4 Y_3 & a_4 Y_3 & a_4 Z_3 & -b_4 & a_4 & 0\\
-b_4 X_4 & -b_4 Z_4 & a_4 X_4 & a_4 Y_4 & a_4 Y_4 & a_4 Z_4 & -b_4 & a_4 & 0\\
   \end{psmallmatrix},\\   
\xx &= 
    \begin{pmatrix}
        R_{11} & R_{12} & R_{13} & R_{21} & R_{22} & R_{23} & T_1 & T_2 & T_3
    \end{pmatrix}^T,\\
\bb    &=
    \begin{pmatrix}
        0 & -b_1 & 0 & 0 & 0 & 0
    \end{pmatrix}^T.
\end{align*}
Solving for translation as in~\eqref{eq:T-backsolve}, we have
\begin{equation}
\begin{split}
    T_1 &= \frac{a_1}{b_1}(-R_{21}X_1 - R_{22}Y_1 - R_{23}Z_1),\\
    T_2 &= -R_{21}X_1 - R_{22}Y_1 - R_{23}Z_1,\\
    T_3 &= -1 + \frac{1}{b_1}(-R_{21}X_1 - R_{22}Y_1 - R_{23}Z_1).
\end{split}
\end{equation}
Furthermore, we express $R_{23}$ from the fourth equation as
\begin{equation}
    R_{23} = \frac{1}{Z_1-Z_2}(R_{21}(X_2-X_1) + R_{22}(Y_2-Y_1)), \label{eq:R23}
\end{equation}
and we solve for $R_{11}$, $R_{12}$ using the last two equations.

Now, we can express $R_{23}$, $T_2$, $T_1$, $R_{12}$, $R_{11}$ as linear combinations of $R_{13}$, $R_{21}$, $R_{22}$ in the following form:
\begin{equation}
    \begin{split}
        R_{23} &= c_1R_{21} + c_2R_{22},\\
        T_{2} &= c_3R_{21} + c_4R_{22},\\
        T_{1} &= c_5R_{21} + c_6R_{22},\\
        R_{12} &= c_7R_{13} + c_8R_{21} + c_9R_{22},\\
        R_{11} &= c_{10}R_{13} + c_{11}R_{21} + c_{12}R_{22}.
    \end{split}\label{eq:linear}
\end{equation}

The non-linear internal constraints imposed on the elements of $R$ have the form:
\begin{equation}
\begin{split}
    R_{11}^2 + R_{12}^2 + R_{13}^2 &= 1,\\
    R_{21}^2 + R_{22}^2 + R_{23}^2 &= 1,\\
    R_{11}R_{21} + R_{12}R_{22} + R_{13}R_{23} &= 0.
\end{split}\label{eq:nonlinear}
\end{equation}
We substitute \eqref{eq:linear} into \eqref{eq:nonlinear}, obtaining equations of the form
\begin{align}
\begin{pmatrix}
d_1 & d_2 & d_3 & d_4 & d_5 & d_6 
\end{pmatrix}
\nu (R)
&=1, \label{eq:first}\\
\begin{pmatrix}
0 & 0 & 0 & d_7 & d_8 & d_9
\end{pmatrix}
\nu (R) &= 1,     \label{eq:second}
\\
\begin{pmatrix}
0 & d_{10} & d_{11} & d_{12} & d_{13} & d_{14} 
\end{pmatrix}
\nu (R) &= 0,     \label{eq:third}
\end{align}
where 
\begin{equation}\label{eq:veronese-vector}
\nu (R) = \begin{pmatrix}
R_{13}^2 & R_{13}R_{21} & R_{13}R_{22} & R_{21}^2 R_{21}R_{22} & R_{22}^2
\end{pmatrix}^T.
\end{equation}
We express $R_{13}$ from \eqref{eq:third} as
\begin{equation}\label{eq:r13}
    R_{13} = -\frac{ d_{12} R_{21}^2 + d_{13} R_{21}R_{22} + d_{14} R_{22}^2 }{ d_{10}R_{21} + d_{11}R_{22} }.
\end{equation}
Substituting~\eqref{eq:r13} into~\eqref{eq:first}, we obtain
\begin{align}
        &d_1\frac{(d_{12} R_{21}^2 + d_{13} R_{21}R_{22} + d_{14} R_{22}^2)^2}{(d_{10}R_{21} + d_{11}R_{22})^2} \, - \nonumber \\ &d_2\frac{d_{12}R_{21}^3 + d_{13}R_{21}^2R_{22} + d_{14}R_{21}R_{22}^2}{d_{10}R_{21} + d_{11}R_{22}} \, - \nonumber \\ &d_3\frac{d_{12}R_{21}^2R_{22} + d_{13}R_{21}R_{22}^2 + d_{14}R_{22}^3}{d_{10}R_{21} + d_{11}R_{22}} \, + \nonumber \\ &d_4R_{21}^2 + d_5R_{21}R_{22} + d_6R_{22}^2 = 1.
\label{eq:first_subs}
\end{align}
Subtracting~\eqref{eq:second} from~\eqref{eq:first_subs},
\begin{align}
&d_1\frac{(d_{12} R_{21}^2 + d_{13} R_{21}R_{22} + d_{14} R_{22}^2)^2}{(d_{10}R_{21} + d_{11}R_{22})^2} \, - \nonumber \\
&d_2\frac{d_{12}R_{21}^3 + d_{13}R_{21}^2R_{22} + d_{14}R_{21}R_{22}^2}{d_{10}R_{21} + d_{11}R_{22}} \, -\nonumber \\ 
&d_3\frac{d_{12}R_{21}^2R_{22} + d_{13}R_{21}R_{22}^2 + d_{14}R_{22}^3}{d_{10}R_{21} + d_{11}R_{22}} \, +  \nonumber \\
&(d_4-d_7)R_{21}^2 + (d_5-d_8)R_{21}R_{22} + (d_6-d_9)R_{22}^2 = 0.\label{eq:subtracted}
\end{align}
We then clear denominators in~\eqref{eq:subtracted}, multiplying by $(d_{10}R_{21} + d_{11}R_{22})^2$ to get a polynomial equation
\begin{equation}
    \begin{split}
        &d_1 (d_{12} R_{21}^2 + d_{13} R_{21}R_{22} + d_{14} R_{22}^2)^2 \, - \\
      \bigg( \Big( &d_2 (d_{12}R_{21}^3 + d_{13}R_{21}^2R_{22} + d_{14}R_{21}R_{22}^2) \, - \\
        &d_3(d_{12}R_{21}^2R_{22} + d_{13}R_{21}R_{22}^2 + d_{14}R_{22}^3) \Big)  \, \cdot  \\
 (&d_{10}R_{21} + d_{11}R_{22})  + ((d_4-d_7)R_{21}^2 + \\
        (&d_5-d_8)R_{21}R_{22} \, + 
        (d_6-d_9)R_{22}^2)\bigg) \cdot \\
        (&d_{10}R_{21} + d_{11}R_{22})= 0.
    \end{split}\label{eq:multiplied}
\end{equation}
Upon expanding equation \eqref{eq:multiplied}, we find that it takes the form
\vspace{-0.5cm}
\begin{equation}
    \alpha_1 R_{21}^4 + \alpha_2 R_{21}^3R_{22} + \alpha_3 R_{21}^2R_{22}^2 + \alpha_4 R_{21}R_{22}^3 + \alpha_5 2R_{22}^4 = 0 \label{eq:multiplied_simple}
\end{equation}
Similarly to~\eqref{eq:v-quadratic}, we divide \eqref{eq:multiplied_simple} by $R_{22}^4$ and define a new variable $v=\frac{R_{21}}{R_{22}}$, and thereby deduce the univariate quartic
\begin{equation}\label{eq:v-quartic}
    \alpha_1 v^4 + \alpha_2 v^3 + \alpha_3 v^2  + \alpha_4 v + \alpha_5 = 0.
\end{equation}
After solving for $v$, we can solve for $R_{21}, R_{22}$ using \eqref{eq:second}, and for the other equations using \eqref{eq:third}, and \eqref{eq:linear}.

To summarize, we provide the outline of steps for solving the P1P2L absolute pose problem in Figure~\ref{alg:p1p2l}.
Algebraically, the nontrivial steps in Figure~\ref{alg:p1p2l} are second and third, which require solving a univariate quartic equation and computing square roots, respectively.

\begin{figure}
\begin{enumerate}
    \item[(1)] Transform data in world and camera frames to the special reference frames satisfying~\eqref{eq:world-pt-transformed},~\eqref{eq:world-ln1-transformed}, \eqref{eq:world-ln2-transformed},~\eqref{eq:C2},~\eqref{eq:cam-pt-transformed},~\eqref{eq:cam-ln1-transformed},~\eqref{eq:cam-ln2-transformed}.
    \item[(2)] Compute up to 4 solutions in $v$ to the quartic~\eqref{eq:v-quartic}.
    \item[(3)] For each root $v$ in step (2), recover $2$ solutions in $\{ R_{2 1}, R_{2 2} \}$ using $R_{2 1} = R_{2 2}v$ and \eqref{eq:second}.
    \item[(4)] Recover remaining coordinates for each solution using~\eqref{eq:third},\eqref{eq:linear}.
    \item[(5)] Reverse the transformation applied in step (1).
    \hrule 
\end{enumerate}
\caption{Generic \textbf{P1P2L solver.} See Sec.~\ref{subsec:p1p2l} for details.}\label{alg:p1p2l}
\end{figure}

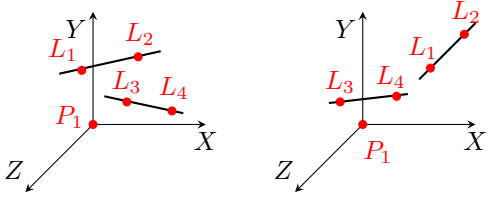
\begin{figure}
    \centering
    \begin{tabular}{c c}
       \begin{tikzpicture}[scale=1.5]

\draw[-stealth] (0,0) -- (0,1);
\draw[-stealth] (0,0) -- (1,0);
\draw[-stealth] (0,0) -- (-0.6,-0.6);

\node at (1,-0.15) {$X$};
\node at (-0.15,0.85) {$Y$};
\node at (-0.7,-0.4) {$Z$};

\filldraw[red] (0.0,0.0) circle (1pt) node[anchor=east,xshift=-0.0cm, yshift=0.1cm]{$P_1$};

\draw[black, thick] (-0.3,0.45) -- (0.6,0.65);
\filldraw[red] (-0.1,0.48) circle (1pt) node[anchor=south,xshift=-0.25cm, yshift=0.0cm]{$L_1$};
\filldraw[red] (0.4,0.6) circle (1pt) node[anchor=west,xshift=-0.3cm, yshift=0.3cm]{$L_2$};

\draw[black, thick] (0.1,0.25) -- (0.8,0.1);
\filldraw[red] (0.3,0.2) circle (1pt) node[anchor=south,xshift=-0.0cm, yshift=0.0cm]{$L_3$};
\filldraw[red] (0.7,0.12) circle (1pt) node[anchor=west,xshift=-0.3cm, yshift=0.3cm]{$L_4$};

\end{tikzpicture}  &  \begin{tikzpicture}[scale=1.5]

\draw[-stealth] (0,0) -- (0,1);
\draw[-stealth] (0,0) -- (1,0);
\draw[-stealth] (0,0) -- (-0.6,-0.6);

\node at (1,-0.15) {$X$};
\node at (-0.15,0.85) {$Y$};
\node at (-0.7,-0.4) {$Z$};

\filldraw[red] (0.0,0.0) circle (1pt) node[anchor=north,xshift=0.2cm, yshift=-0.1cm]{$P_1$};

\draw[black, thick] (0.5,0.4) -- (1.0,0.9);
\filldraw[red] (0.6,0.5) circle (1pt) node[anchor=south,xshift=-0.1cm, yshift=0.0cm]{$L_1$};
\filldraw[red] (0.9,0.8) circle (1pt) node[anchor=west,xshift=-0.3cm, yshift=0.3cm]{$L_2$};

\draw[black, thick] (-0.3,0.19) -- (0.4,0.27);
\filldraw[red] (-0.2,0.2) circle (1pt) node[anchor=south,xshift=-0.0cm, yshift=0.0cm]{$L_3$};
\filldraw[red] (0.3,0.25) circle (1pt) node[anchor=west,xshift=-0.4cm, yshift=0.26cm]{$L_4$};

\end{tikzpicture}
    \end{tabular}
    \vspace{-0.2cm}
    \caption{Stabilization of P1P2L solver. The frame is rotated to align the line $\overline{L_1 L_2}$ with the z-axis. See Sec.~\ref{subsec:stabilization} for details.}
    \label{fig:p1p2l_stabilization}
\end{figure}

\subsection{Stabilizing the P1P2L Solver}\label{subsec:stabilization}
%\begin{itemize}
%    \color{red}
%    \item Distinguish between degenerate cases of the actual problem vs the algorithm used to solve it
%    \item Describe the source of instability that occurs when the last coordinate of X2-X1 (or something like that) is equal to zero $Z_1-Z_2$
%    \item Propose rotating the world frame such that the value is maximized (==align this with the z-axis)
%    \item Add an figure illustrating this
%    \item Show the synthetic tests that prove that this simple fix improves the stability quite significantly 
%    
%\end{itemize}
In this section, we outline a method to increase the stability of the P1P2L solver (Sec.~\ref{subsec:p1p2l}).

Our proposed P1P2L solution faces a degeneracy when $Z_1 - Z_2 = 0$, since this leads to a division by zero in equation~\eqref{eq:R23}. Furthermore, our observations indicate that the result of the solver is unstable if the value of $Z_1 - Z_2$ is close to zero. Note, that this instability is specific to our solver and not inherent to the P1P2L problem.

This degeneracy may be interpreted geometrically as follows: the values of $Z_1$ and $Z_2$ are the last coordinates of the points $L_1$, $L_2$, which span the first 3D line. Therefore, the vector $L_1 - L_2$ represents the direction of the line, and $Z_1 - Z_2$ represents the last coordinate of this direction.

Since the special reference frame used in~\Cref{subsec:p1p2l} is independent of the rotation of the world frame, we can remove the source of instability by rotating the world frame such that the first line $\overline{L_1 L_2}$ aligns with the z-axis. See~\Cref{fig:p1p2l_stabilization} for an illustration. 

\begin{table}[]
    \centering
    \begin{tabular}{ccccc}
    %%\hline
    %Method & Mean R & Median R & Max R  & Mean T & Median T & Max T \\
    %%Method & Mean R & Med. R  & Mean T & Med. T \\
    %%\hline

    %%Ours no fix & 0.00029 & 1.0e-14 & 0.00051 & 1.7e-13 \\
    %%Ours + fix & \textbf{1.2e-07} & \textbf{4.4e-15} & \textbf{2.0e-06} & \textbf{7.1e-14}\\
    %%\hline
    %%3Q3 no fix & 0.00017 & 6.8e-15 & 0.00011 & 1.4e-13 \\
    %%3Q3 + fix & 3.3e-05 & 7.2e-15 & 3.4e-05 & 1.5e-13 \\

    \hline
    \textbf{Generic} & Mean R & Med. R  & Mean T & Med. T \\
    \hline
    P1P2L no fix & 0.00050 & 1.0e-14 & 0.00065 & 1.8e-13 \\
    P1P2L fix & \textbf{9.0e-09} & \textbf{4.2e-15} & \textbf{3.4e-07} & \textbf{7.0e-14} \\
    %P1P2L 3Q3 & 6.2e-05 & 6.6e-15 & 9.0e-05 & 1.3e-13\\
    \hline
    P2P1L no fix & \textbf{7.1e-11} & \textbf{1.4e-15} & \textbf{1.2e-09} & \textbf{2.1e-14} \\
    P2P1L fix & 3.2e-09 & 2.6e-15 & 7.4e-08 & 4.3e-14 \\
    %P2P1L 3Q3 & 2.6e-05 & 6.5e-15 & 1.1e-05 & 1.2e-13 \\
    \hline
    \hline
    \textbf{Coplanar} & Mean R & Med. R  & Mean T & Med. T \\
    \hline
    P1P2L no fix & \textbf{0.00022} & \textbf{9.6e-15} & \textbf{0.00030} & \textbf{1.75e-13} \\
    P1P2L fix & 1.1 & 0.79 & 1.2 & 0.96 \\
    %P1P2L 3Q3 & \textbf{9.3e-05} & 1.2e-14 & \textbf{5.0e-05} & 3.7e-13 \\
    \hline
    P2P1L no fix & 2.4 & 3.14 & 2.2 & 3.14  \\
    P2P1L fix & \textbf{1.2e-12} & \textbf{4.0e-15} & \textbf{7.9e-11} & \textbf{6.3e-14} \\
    %P2P1L 3Q3 & 4.1e-05 & 2.0e-14 & 2.1e-05 & 5.2e-13 \\
    \hline

    %P1P2L Ours no fix & 0.00029 & 1.0e-14 & 3.14 & 0.00051 & 1.7e-13 & 3.14 \\
    %%P1P2L Ours fix 1 & 7.3e-07 & 6.0e-15 & 0.067 & 2.5e-06 & 9.9e-14 & \textbf{0.11}\\
    %P1P2L Ours fix 2 & \textbf{1.2e-07} & \textbf{4.4e-15} & \textbf{0.010} & \textbf{2.0e-06} & \textbf{7.1e-14} & 0.13\\
    %\hline
    %P1P2L 3Q3 & 0.00017 & 6.8e-15 & 2.63 & 0.00011 & 1.4e-13 & 1.12 \\
    %%P1P2L 3Q3 + fix 1 & 8.5e-05 & 6.6e-15 & 2.49 & 7.0e-05 & 1.4e-13 & 1.12 \\
    %P1P2L 3Q3 + fix 2 & 3.3e-05 & 7.2e-15 & 2.60 & 3.4e-05 & 1.5e-13 & 1.01 \\

    %P1P2L Ours no fix & 0.0002937370641756124 & 1.0282700000000001e-14 & 3.14 & 0.0005076983931440067 & 1.749955e-13 & 3.14 \\
    %P1P2L Ours fix 1 & 7.309754991496699e-07 & 6.008515e-15 & 0.0670172 & 2.5471503866712763e-06 & 9.9124e-14 & \textbf{0.109648}\\
    %P1P2L Ours fix 2 & \textbf{1.1997876558785292e-07} & \textbf{4.3518900000000006e-15} & \textbf{0.0102093} & \textbf{1.9654832049072605e-06} & \textbf{7.10601e-14} & 0.131242\\
    %\hline
    %P1P2L \textit{Ramalingam} & 0.00016924435172545888 & 6.812925e-15 & 2.63091 & 0.00011289182932821801 & 1.399025e-13 & 1.12129 \\
    %P1P2L \textit{Ramalingam} + fix 1 & 8.532122757401411e-05 & 6.58511e-15 & 2.48821 & 7.001903392996162e-05 & 1.36824e-13 & 1.10523 \\
    %P1P2L \textit{Ramalingam} + fix 2 & 3.347808528353049e-05 & 7.24604e-15 & 2.59985 & 3.365500050633916e-05 & 1.5025749999999998e-13 & 1.00981 \\
    \hline
    \end{tabular}
    \caption{The mean and median of the rot.(R) and tran.(T) errors over $1e5$ noiseless \textbf{generic} (\textit{top}) and \textbf{coplanar} (\textit{bottom}) samples. In radians. The best results are marked bold.}
    %\caption{\textbf{Fixing the instability in the P1P2L solver.} The mean, median and max values of the errors. The best results are marked bold. See Sec.~\ref{subsec:stabilization} for details.}
    \label{tab:fixing_instability}
\end{table}

%TODO move the table here
%TODO also create a figure of this

\subsection{Resolving the coplanar case}\label{subsec:coplanar}

In this section, we discuss the performance of the P2P1L (Sec.~\ref{subsec:p2p1l}) and P1P2L (Sec.~\ref{subsec:p1p2l}) solvers in the case, where all 3D points and 3D lines are coplanar.

\noindent
\textbf{P2P1L.} The P2P1L solver presented in~\cref{subsec:p2p1l} (\textit{P2P1L no fix}) is degenerate in coplanar situations, since then $Z_4 = 0$ \eqref{eq:world-ln-transformed}, causing a division by zero issue in equation~\eqref{eq:linear_constraints-p2p1l}. To resolve this issue, we propose a modified solver (\textit{P2P1L fix}), which handles the coplanar case. See SM~\ref{sm:p2p1l-coplanar} for a description of this modified solver.

\noindent
\textbf{P1P2L.} The original version of the P1P2L solver (\textit{P1P2L no fix},~\cref{subsec:p1p2l}) is able to handle the coplanar case. However, the stabilized P1P2L solver (\textit{P1P2L fix},~\cref{subsec:stabilization}) fails due to a degeneracy for coplanar input.

We conducted a synthetic experiment to evaluate the impact of the proposed stablization scheme.
In this experiment, we used the same setting as in~\Cref{subsec:synth_experiments}.
%The parameters of this experiment were set according to~\Cref{subsec:synth_experiments}.
%
The results, presented in~\Cref{tab:fixing_instability} demonstrate that solvers \textit{P2P1L no fix} and \textit{P1P2L fix} achieve superior numerical stability in the generic case, while solvers \textit{P2P1L fix} and \textit{P1P2L no fix} handle the coplanar case. Since it is simple to detect the coplanar case, we recommend to use solvers \textit{P2P1L fix} and \textit{P1P2L no fix} in the coplanar case and solvers \textit{P2P1L no fix} and \textit{P1P2L fix} in the generic case.
%rotating the world frame enhances the numerical stability of the solver.

%TODO basically repeat the answer from rebuttal

%TODO create supplementary and add there the details about the P2P1L fix solver

%TODO move the experiment (last paragraph of previous section here) and add there the coplanar case from the rebuttal

%TODO create some flowchart to show what solver to use

\section{Experiments}\label{sec:experiments}

%\begin{itemize}
%    \color{red}
%    \item In this section, we experimentally compare the proposed solvers with the state-of-the-art methods.
%    \item All the solvers are implemented in C++ and the experiments are conducted on a desktop computer with AMD Ryzen 9 with 3.9 GHz.
%    \item We compare with the following solvers: 3Q3, Ramalingam. For 3Q3 we use the publicly available implementation from Poselib. Since the implementation of Ramalingam is not publicly available, we use our own implementation. Since it is not directly mentioned, which method should be used to solve the linear equations, we compare with 2 variants: 1. SVD, 2. LU.
%    \item In section ..., we provide the analysis of the numerical stability and runtime of the solvers on synthetic data. In section ..., we show an evaluation of the solvers within the RANSAC scheme.
%\end{itemize}

In this section, we experimentally compare the proposed solvers with the state-of-the-art methods, specifically 3Q3 \cite{DBLP:conf/cvpr/KukelovaHF16} and Ramalingam \cite{DBLP:conf/icra/RamalingamBS11}. For 3Q3, we employ the publicly available implementation from Poselib. As the implementation of Ramalingam is not publicly accessible, we have created our own implementation. Since it is not explicitly specified which method should be used for solving the linear equations, we compare with two variants: SVD and LU decomposition. All the solvers have been implemented in C++, and the experiments are conducted on a desktop computer with an AMD Ryzen 9 CPU with 3.9 GHz.

In Sec.~\ref{subsec:synth_experiments}, we provide an analysis of the numerical stability and runtime performance of the solvers on synthetic data. Subsequently, in Section \ref{subsec:ransac_experiments}, we present an evaluation of the solvers within the RANSAC scheme.

\begin{figure}[!]
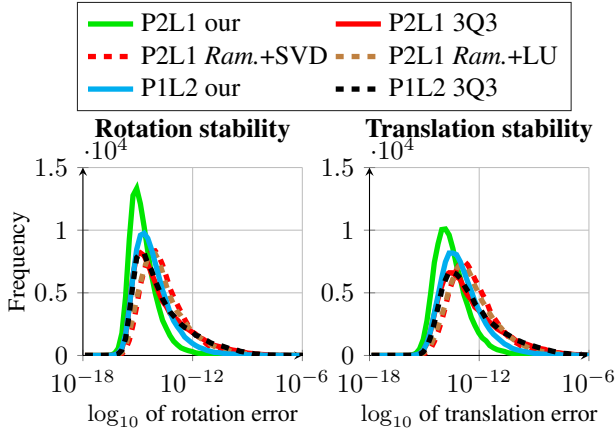

    \centering
    \begin{tikzpicture}

\input{plot/rotation_stability}
\input{plot/translation_stability}

\end{tikzpicture}
    \vspace{-0.7cm}
    \caption{\textbf{Stability test.} Histogram of rotation (\textit{l.}), and translation (\textit{r.}) errors computed over $1e5$ noiseless samples.}
    \label{fig:stability}
\end{figure}

\begin{table}[]
    \centering
    \begin{tabular}{r r r r r r}
    \hline
    Method & Mean & Med. & Min & Max \\
    \hline
    P2P1L Ours & \textbf{313.8} & \textbf{324.9} & \textbf{230.8} & \textbf{3061.0}
  \\
    P2P1L 3Q3 & 1860.6 & 1909.9 & 1439.1 & 10102.1  \\
    P2P1L \textit{R.}+S & 8897.5 & 9491.2 & 5805.1 & 49984.1 \\
    P2P1L \textit{R.}+L & 4720.8 & 5239.5 & 2763.3 & 15362.6 \\
    \hline
    P1P2L Ours & \textbf{504.0} & \textbf{521.0} & \textbf{364.0} & \textbf{4554.4} \\
    P1P2L 3Q3 & 1967.1 & 2008.2 & 1483.9 & 12931.1  \\
    \hline
    \end{tabular}
    \caption{\textbf{Runtime} The mean, median and max values of the runtime. In nanoseconds. Best values for each problem are bold.}
    \label{tab:runtime}
\end{table}

\begin{table*}[!htb]
    \centering
    \begin{tabular}{c c c c c c c}
    \hline
    Method & Mean R & Median R & Max R  & Mean T & Median T & Max T \\
    \hline
    P2P1L Ours & \textbf{5.3e-12} & \textbf{1.4e-15} & \textbf{1.2e-07} & \textbf{3.7e-10} & \textbf{2.1e-14} & \textbf{2.2e-05} \\
    P2P1L 3Q3 & 2.8e-05 & 6.5e-15 & 2.7 & 2.0e-05 & 1.2e-13 & 1.04 \\
    P2P1L \textit{Ramalingam} (SVD) & 4.9e-09 & 1.5e-14 & 0.00041 & 1.6e-07 & 2.7e-13 & 0.0099 \\
    P2P1L \textit{Ramalingam} (LU) & 4.7e-07 & 1.3e-14 & 0.040 & 2.3e-05 & 2.4e-13 & 0.10
 \\
    \hline
    PP1P2L Ours (fix 2) & \textbf{1.2e-07} & \textbf{4.4e-15} & \textbf{0.010} & \textbf{2.0e-06} & \textbf{7.1e-14} & \textbf{0.13}\\
    P1P2L 3Q3 & 3.3e-05 & 7.2e-15 & 2.60 & 3.4e-05 & 1.5e-13 & 1.01 \\
    \hline
    \end{tabular}
    \vspace{-0.2cm}
    \caption{The mean, median and max values of the rot.(R) and tran.(T) errors over $1e5$ noiseless samples. The best results are marked bold.}
    \label{tab:error}
\end{table*}

\subsection{Synthetic experiments}\label{subsec:synth_experiments}

%\begin{itemize}
    %\color{red}
    %\item Give an initializing sentence about what is happening in this section.
    %\item Describe how the data is generated: 1. how do we generate R,t; 2. how do we generate the 3D points/lines, 3. how do we obtain the 2D points/lines
    %\item Describe how the errors are calculated
    %\item Point to the plot and the table: (In Figure~\ref{fig:stability}, we show the histograms of the rotation %and translation errors of all considered solvers.)
    %\item Say that all solvers are numerically stable. Say that our are more numerically stable.
    %\item All minimal solvers are equal. But some of them are more equal than others.
    %\item Point to the table: In Table~\ref{tab:error}, we show the ...
    %\item Describe what happens: note that ...
    %\item Point to the time evaluation
%\end{itemize}

In this section, we present an analysis of the numerical stability and runtime performance of the minimal solvers on synthetic data.

We generate instances of each minimal problem (either P2P1L or P1P2L), according to the following procedure:

\begin{itemize}
    \item \textbf{Rotation Matrix} ($R$): An axis $v \in \mathbb{R}^3$ is sampled from the uniform distribution on the unit sphere, and an angle $\alpha$ is sampled from the normal distribution $\mathcal{N}(0, 1).$ The rotation matrix $R$ is then constructed using the angle-axis formula, $R = I + \sin (\alpha ) [v]_\times + (1-\cos (\alpha )) [v]_\times^2 $.
    \item \textbf{Translation} ($T$): The camera center $C$ is sampled uniformly at random from the unit sphere. The translation vector $T$ is computed as $T = -R C$.
    \item \textbf{Point correspondence}: A 3D point $X_i$ is sampled from the trivariate normal distribution with mean vector $\mu = [0, 0, 5]^T$ and standard deviation $\sigma = 1$ in each component. The corresponding 2D point $D_i$ is obtained by projecting $X_i$ onto a pinhole camera with the pose $(R, T)$.
    \item \textbf{Line correspondence}: Two 3D points, $L_i$ and $L_{i+1}$, are sampled as described in the previous bullet-point. These points define the 3D line $L$. Two 2D points, $D_i$ and $D_{i+1}$, are obtained by sampling two points on the 3D line $L$ and projecting them onto the camera with the pose $(R, T)$.
\end{itemize}

In the case of P2P1L, two points and one line are sampled, while in the case of P1P2L, one point and two lines are sampled. Let $(R_{est}, T_{est})$ be the pose obtained by the minimal solver. We measure the rotation error as the angle $\xi_R = \arccos( (1+ \operatorname{trace} (R_{est}^T R)) / 2)$. We calculate the translation error $\xi_T$ as $\xi_T = \frac{|T_{est} - T|}{|T|}$. The histograms of rotation errors ($\xi_R$) and translation errors ($\xi_T$) for all considered solvers are depicted in Figure~\ref{fig:stability}. Furthermore, summary statistics including the median, mean, and maximum errors are provided in Table~\ref{tab:error}. The results indicate that all minimal solvers are stable. Our algebraically optimal solvers demonstrate superior stability compared to the previous solvers.

The runtime evaluation for all considered solvers is shown in Table.~\ref{tab:runtime}. Our P2P1L solver requires $313.8 ns$ on average, which is about 6x faster than the 3Q3 solver \cite{DBLP:conf/cvpr/KukelovaHF16}. Similarly, our P1P2L solver requires on average $504.0 ns$, which is about 4x faster than the 3Q3 solver. These results show a significant speedup achieved by our solvers in comparison to the previous methods.

% Let us describe how to generate one instance of the minimal problem (P2P1L or P1P2L). To generate the rotation matrix $R$, we first sample the axis $v \in \RR^3$ from the uniform distribution on the sphere, and the angle $\alpha \in \RR$ from the normal distribution N(0,1). Then, we construct the rotation $R$ with angle $\alpha$ and axis $\v$. We sample the camera center $C \in \RR^3$ from the three-dimensional normalized normal distribution, and construct the translation as $T = -R*C$.  A 3D point $X_i \in \RR^3$ is sampled from the normal distribution with $\mu = [0 0 5]^T$ and $\sigma = 1$ and its corresponding 2D point $D_i$ is obtained by projecting $X_i$ onto the pinhole camera with pose $(R, T)$. To generate a 3D line $L$, we sample 2 3D points $L_i, L_{i+1}$ according to the same principle as before. These points span the line $L$. To obtain the 2D points $D_i, D_{i+1}$ that span the 2D line $l$, we sample two points on the 3D line and project them to the camera with pose $(R, T)$. In the case of P2P1L, we sample 2 points and 1 line, and in the case of P1P2L, we sample 1 point and 2 lines.

%Let $(R_{est}, T_{est})$ be the pose obtained by the minimal solver. We measure the rotation error $\xi_R$ as the angle of the rotation $R_{est}^T R$ and the translation error $\xi_t$ as $\xi_t = \frac{}{}$.

\begin{table*}[]
    \centering
    \begin{tabular}{r | r r r r | r r }
         %& Model House & Corridor & Merton I & Merton II & Merton III & Library & \\
          & \multicolumn{4}{|c|}{P2P1L} & \multicolumn{2}{c}{P1P2L} \\
         Dataset & OUR & 3Q3 & \textit{Ram.} SVD & \textit{Ram.} LU & OUR & 3Q3 \\
         \midrule
         Model House & \textbf{7.72} (0.85x) & 9.09 (1x) & 20.69 (2.28x) & 17.30 (1.90x) & \textbf{8.06} (0.85x) & 9.52 (1x) \\
         Corridor & \textbf{11.32} (0.91x) & 12.46 (1x) & 26.56 (2.13x) & 23.54 (1.89x) & \textbf{13.3} (0.90x) & 14.76 (1x) \\
         Merton I & \textbf{35.93} (0.98x) & 36.83 (1x) & 77.96 (2.12x) & 74.87 (2.03x) & \textbf{28.85} (0.64x) & 45.41 (1x) \\
         Merton II & \textbf{33.30} (0.97x) & 34.27 (1x) & 74.37 (2.17x) & 71.35 (2.08x) & \textbf{26.7} (0.67x) & 39.64 (1x) \\
         Merton III & \textbf{24.04} (0.96x) & 25.07 (1x) & 52.07 (2.08x) & 48.97 (1.95x) & \textbf{10.91} (0.37x) & 29.67 (1x) \\
         Library & \textbf{32.42} (0.97x) & 33.31 (1x) & 69.29 (2.08x) & 66.23 (1.99x) & \textbf{10.04} (0.28x) & 35.98 (1x) \\
         Wadham & \textbf{39.60} (0.98x) & 40.51 (1x) & 86.10 (2.13x) & 83.00 (2.05x) & \textbf{22.96} (0.45x) & 50.79 (1x) \\
         \midrule
         Avg. Speed-up & 0.94x\phantom{)} & 1x\phantom{)} & 2.14x\phantom{)} & 1.99x\phantom{)} & 0.59x\phantom{)} & 1x\phantom{)} \\
    \end{tabular}
    \vspace{-0.1cm}
    \caption{\textbf{RANSAC timing}, on Oxford Multi-view dataset \cite{oxdata_multiview}, in milliseconds. Speed-up compared to the 3Q3 in the bracket.}
    \label{tab:real_test_time}
\end{table*}

%RANSAC ERROR TABLE WAS HERE

\begin{figure}[!]
    \centering
    \begin{tabular}{cc}
       \includegraphics[width=8.7em]{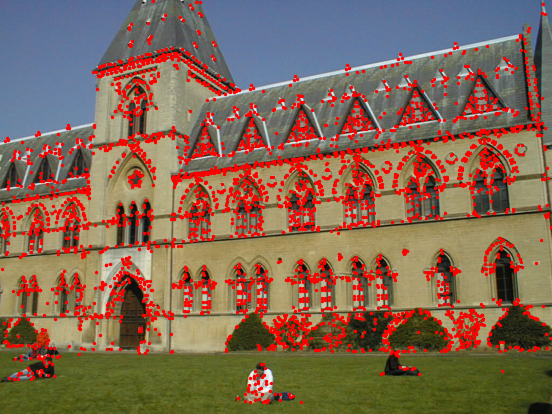}  & 
       \includegraphics[width=8.7em]{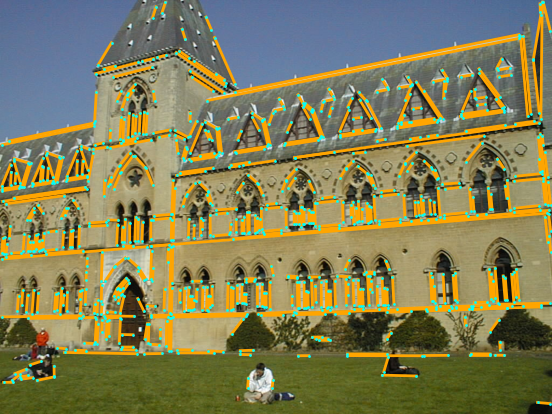} \\
       \includegraphics[width=8.7em]{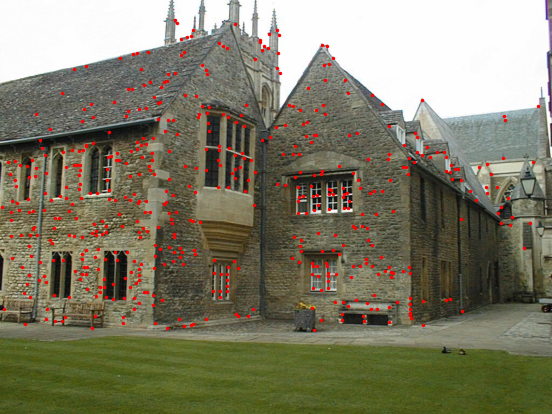}  & 
       \includegraphics[width=8.7em]{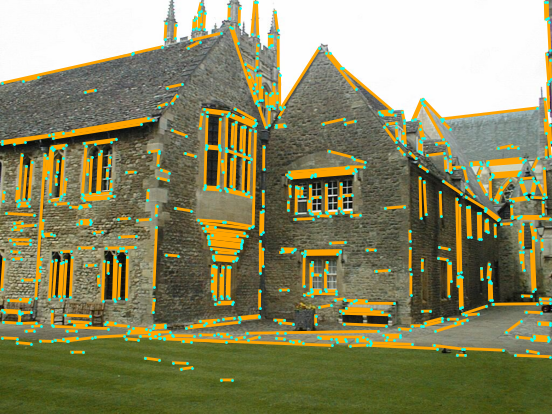} 
    \end{tabular}
    \vspace{-0.3cm}
    \caption{Example images of University Library~(\textit{top}) and Merton College III~(\textit{bottom}), from~\cite{oxdata_multiview}.}
    \label{fig:library-lines}
\end{figure}
%\vspace{-0.1cm}

\subsection{Experiments in RANSAC}\label{subsec:ransac_experiments} 

Here, we provide an analysis of the minimal solvers within the RANSAC scheme \cite{DBLP:conf/ijcai/BollesF81}, using the Oxford Multi-view Data~\cite{oxdata_multiview}. This dataset includes a variety of indoor and outdoor scenes, some of which contain both point and line matches. An example of images from this dataset is illustrated in~\Cref{fig:library-lines}.

%We use the Poselib \cite{} implementation of the LO RANSAC \cite{}. In this implementation, the local optimization is applied every time the obtained solution is better than the so-far-best one. We use the following RANSAC parameters: maximum number of iterations = 100000, minimum number of iterations = 1000, success probability = 0.9999, inlier threshold = 1px. The random seed is fixed to ensure a fair comparison between the solvers.
%\vspace{-0.6cm}
We utilize the Poselib~\cite{PoseLib} implementation of LO-RANSAC~\cite{DBLP:conf/dagm/ChumMK03}. In this implementation, local optimization is applied both when the new solution surpasses the currently best one and at the end of the entire procedure. The RANSAC parameters used include a maximum number of iterations set to 100000, a minimum number of iterations set to 1000, a success probability of 0.9999, and an inlier threshold of 1 pixel. To ensure a fair comparison between the solvers, we use a fixed random seed.
%\vspace{-0.07cm}

%Since all the solvers are stable, we do not expect any significant difference in the pose estimated using different solvers. Thus, we mostly concentrate on comparing the runtime of the RANSAC scheme. We show the runtime comparison in Table.~\ref{tab:real_test_time}. This table shows an average runtime for every scene from the Oxford multiview dataset and for every considered solver. Additionaly, it also shows a speedup compared to the 3Q3 solver. The results indicate that our solvers are always faster than the 3Q3 \cite{} and Ramalingam \cite{} solvers. However, there is a large variance among the scenes that ranges between 0.85 and 0.98 for the P2P1L problem and between 0.28 and 0.9 for the P1P2L solver. This difference can be explained by the fact that for different images, RANSAC spends different portion of time on scoring the models and on local optimization. The larger portion of time is spent on these tasks, the less significant speed-up we can reach. 
Since all the solvers are stable, we do not expect any significant difference in the estimated pose when using different solvers. Therefore, our primary focus is on comparing the runtime of the RANSAC scheme. The runtime comparison is presented in Table~\ref{tab:real_test_time}, which displays the average runtime for each scene in the Oxford multiview dataset and for each considered solver. Additionally, the table includes a speedup value compared to the 3Q3 solver of~\cite{PoseLib}. The results show that our solvers consistently outperform the 3Q3 \cite{DBLP:conf/cvpr/KukelovaHF16} and Ramalingam \cite{DBLP:conf/icra/RamalingamBS11} solvers in terms of runtime. However, there is a notable variation in the speed-up among the scenes, which ranges from 0.85 to 0.98 for the P2P1L problem and from 0.28 to 0.90 for the P1P2L solver. This variation can be attributed to the fact that different scenes require varying proportions of time spent by RANSAC on scoring hypothesized models and local optimization. The more time is dedicated to these tasks, the less significant speedup can be reached.
In cases with fewer matches or a lower inlier ratio, more substantial speedups are possible.

%The error of the estimated pose is shown in Table~\ref{TODO}. As expected, the errors are the same for all solvers. This means, that we are able to reach better runtimes without sacrificing any of the precision.
%Table~\ref{tab:ransac_error} shows the final pose estimation errors in RANSAC. 
The final pose estimation errors in RANSAC are shown in the Supplementary. As expected, the errors are the same for all solvers. This demonstrates that we can achieve improved runtimes without sacrificing any accuracy.

%TODO 

%\vspace{-0.3cm}
\section{Conclusion}
\vspace{-0.1cm}
\label{sec:conclusion}

In revisiting absolute pose from mixed point-line corresopondences, we have developed new solvers, which have the merit of being both algebraically optimal and elementary in nature. 
Moreover, our solvers outperform the state-of-the-art in terms of both runtime and numerical stability. 
Despite the simplicity of many minimal absolute pose problems, our findings suggest that their continued investigation remains worthwhile. The code is available at \url{https://github.com/petrhruby97/efficient\_absolute}
%We will make the code publicly available.
%The code is publicly available.

\textbf{Acknowledgements:} TD was supported by NSF DMS 2103310.
\clearpage 
{
    \small
    \bibliographystyle{ieeenat_fullname}
    \bibliography{main}

\begin{thebibliography}{30}
\providecommand{\natexlab}[1]{#1}
\providecommand{\url}[1]{\texttt{#1}}
\expandafter\ifx\csname urlstyle\endcsname\relax
  \providecommand{\doi}[1]{doi: #1}\else
  \providecommand{\doi}{doi: \begingroup \urlstyle{rm}\Url}\fi

\bibitem[oxd()]{oxdata_multiview}
Multi-view data, oxford visual geometry group.
\newblock https://www.robots.ox.ac.uk/~vgg/data/mview/.
\newblock Accessed: Nov 8 2023.

\bibitem[Agostinho et~al.(2023)Agostinho, Gomes, and Del~Bue]{agostinho2023cvxpnpl}
S{\'e}rgio Agostinho, Jo{\~a}o Gomes, and Alessio Del~Bue.
\newblock {CvxPnPL}: A unified convex solution to the absolute pose estimation problem from point and line correspondences.
\newblock \emph{Journal of Mathematical Imaging and Vision}, 65\penalty0 (3):\penalty0 492--512, 2023.

\bibitem[Barath and Matas(2022)]{DBLP:journals/pami/BarathM22}
Daniel Barath and Jiri Matas.
\newblock Graph-cut {RANSAC:} local optimization on spatially coherent structures.
\newblock \emph{{IEEE} Trans. Pattern Anal. Mach. Intell.}, 44\penalty0 (9):\penalty0 4961--4974, 2022.

\bibitem[Bolles and Fischler(1981)]{DBLP:conf/ijcai/BollesF81}
Robert~C. Bolles and Martin~A. Fischler.
\newblock A ransac-based approach to model fitting and its application to finding cylinders in range data.
\newblock In \emph{Proceedings of the 7th International Joint Conference on Artificial Intelligence, {IJCAI} '81, Vancouver, BC, Canada, August 24-28, 1981}, pages 637--643. William Kaufmann, 1981.

\bibitem[Chen(1991)]{DBLP:journals/pami/Chen91}
Homer~H. Chen.
\newblock Pose determination from line-to-plane correspondences: Existence condition and closed-form solutions.
\newblock \emph{{IEEE} Trans. Pattern Anal. Mach. Intell.}, 13\penalty0 (6):\penalty0 530--541, 1991.

\bibitem[Chum et~al.(2003)Chum, Matas, and Kittler]{DBLP:conf/dagm/ChumMK03}
Ondrej Chum, Jiri Matas, and Josef Kittler.
\newblock Locally optimized {RANSAC}.
\newblock In \emph{Pattern Recognition, 25th {DAGM} Symposium, Magdeburg, Germany, September 10-12, 2003, Proceedings}, pages 236--243. Springer, 2003.

\bibitem[Dhome et~al.(1989)Dhome, Richetin, Laprest{\'{e}}, and Rives]{DBLP:journals/pami/DhomeRLR89}
Michel Dhome, Marc Richetin, Jean{-}Thierry Laprest{\'{e}}, and G{\'{e}}rard Rives.
\newblock Determination of the attitude of 3d objects from a single perspective view.
\newblock \emph{{IEEE} Trans. Pattern Anal. Mach. Intell.}, 11\penalty0 (12):\penalty0 1265--1278, 1989.

\bibitem[Ding et~al.(2023)Ding, Yang, Larsson, Olsson, and {\AA}str{\"{o}}m]{DBLP:conf/cvpr/0001YLOA23}
Yaqing Ding, Jian Yang, Viktor Larsson, Carl Olsson, and Kalle {\AA}str{\"{o}}m.
\newblock Revisiting the {P3P} problem.
\newblock In \emph{{IEEE/CVF} Conference on Computer Vision and Pattern Recognition, {CVPR} 2023, Vancouver, BC, Canada, June 17-24, 2023}, pages 4872--4880. {IEEE}, 2023.

\bibitem[Duff et~al.(2022)Duff, Korotynskiy, Pajdla, and Regan]{galvis1}
Timothy Duff, Viktor Korotynskiy, Tomas Pajdla, and Margaret~H. Regan.
\newblock Galois/monodromy groups for decomposing minimal problems in 3d reconstruction.
\newblock \emph{SIAM Journal on Applied Algebra and Geometry}, 6\penalty0 (4):\penalty0 740--772, 2022.

\bibitem[Fabbri et~al.(2021)Fabbri, Giblin, and Kimia]{DBLP:journals/pami/FabbriGK21}
Ricardo Fabbri, Peter~J. Giblin, and Benjamin~B. Kimia.
\newblock Camera pose estimation using first-order curve differential geometry.
\newblock \emph{{IEEE} Trans. Pattern Anal. Mach. Intell.}, 43\penalty0 (10):\penalty0 3321--3332, 2021.

\bibitem[Fischler and Bolles(1981)]{DBLP:journals/cacm/FischlerB81}
Martin~A. Fischler and Robert~C. Bolles.
\newblock Random sample consensus: {A} paradigm for model fitting with applications to image analysis and automated cartography.
\newblock \emph{Commun. {ACM}}, 24\penalty0 (6):\penalty0 381--395, 1981.

\bibitem[H{\"{a}}ne et~al.(2017)H{\"{a}}ne, Heng, Lee, Fraundorfer, Furgale, Sattler, and Pollefeys]{DBLP:journals/ivc/HaneHLFFSP17}
Christian H{\"{a}}ne, Lionel Heng, Gim~Hee Lee, Friedrich Fraundorfer, Paul Furgale, Torsten Sattler, and Marc Pollefeys.
\newblock 3d visual perception for self-driving cars using a multi-camera system: Calibration, mapping, localization, and obstacle detection.
\newblock \emph{Image Vis. Comput.}, 68:\penalty0 14--27, 2017.

\bibitem[Haralick et~al.(1994)Haralick, Lee, Ottenberg, and N{\"{o}}lle]{DBLP:journals/ijcv/HaralickLON94}
Robert~M. Haralick, Chung{-}Nan Lee, Karsten Ottenberg, and Michael N{\"{o}}lle.
\newblock Review and analysis of solutions of the three point perspective pose estimation problem.
\newblock \emph{Int. J. Comput. Vis.}, 13\penalty0 (3):\penalty0 331--356, 1994.

\bibitem[Hruby et~al.(2023)Hruby, Korotynskiy, Duff, Oeding, Pollefeys, Pajdla, and Larsson]{DBLP:conf/cvpr/HrubyKDOPPL23}
Petr Hruby, Viktor Korotynskiy, Timothy Duff, Luke Oeding, Marc Pollefeys, Tom{\'{a}}s Pajdla, and Viktor Larsson.
\newblock Four-view geometry with unknown radial distortion.
\newblock In \emph{{IEEE/CVF} Conference on Computer Vision and Pattern Recognition, {CVPR} 2023, Vancouver, BC, Canada, June 17-24, 2023}, pages 8990--9000. {IEEE}, 2023.

\bibitem[Hu et~al.(2023)Hu, Zhu, Wang, Yang, and Li]{DBLP:journals/access/HuZWYL23}
Xiaomei Hu, Luying Zhu, Ping Wang, Haili Yang, and Xuan Li.
\newblock Improved {ORB-SLAM2} mobile robot vision algorithm based on multiple feature fusion.
\newblock \emph{{IEEE} Access}, 11:\penalty0 100659--100671, 2023.

\bibitem[Kneip et~al.(2011)Kneip, Scaramuzza, and Siegwart]{DBLP:conf/cvpr/KneipSS11}
Laurent Kneip, Davide Scaramuzza, and Roland Siegwart.
\newblock A novel parametrization of the perspective-three-point problem for a direct computation of absolute camera position and orientation.
\newblock In \emph{The 24th {IEEE} Conference on Computer Vision and Pattern Recognition, {CVPR} 2011, Colorado Springs, CO, USA, 20-25 June 2011}, pages 2969--2976. {IEEE} Computer Society, 2011.

\bibitem[Kukelova et~al.(2016)Kukelova, Heller, and Fitzgibbon]{DBLP:conf/cvpr/KukelovaHF16}
Zuzana Kukelova, Jan Heller, and Andrew~W. Fitzgibbon.
\newblock Efficient intersection of three quadrics and applications in computer vision.
\newblock In \emph{2016 {IEEE} Conference on Computer Vision and Pattern Recognition, {CVPR} 2016, Las Vegas, NV, USA, June 27-30, 2016}, pages 1799--1808. {IEEE} Computer Society, 2016.

\bibitem[Larsson and contributors(2020)]{PoseLib}
Viktor Larsson and contributors.
\newblock {PoseLib - Minimal Solvers for Camera Pose Estimation}, 2020.

\bibitem[Liu et~al.(2023)Liu, Yu, Pautrat, Pollefeys, and Larsson]{DBLP:conf/cvpr/LiuYPPL23}
Shaohui Liu, Yifan Yu, R{\'{e}}mi Pautrat, Marc Pollefeys, and Viktor Larsson.
\newblock 3d line mapping revisited.
\newblock In \emph{{IEEE/CVF} Conference on Computer Vision and Pattern Recognition, {CVPR} 2023, Vancouver, BC, Canada, June 17-24, 2023}, pages 21445--21455. {IEEE}, 2023.

\bibitem[Nist{\'{e}}r et~al.(2007)Nist{\'{e}}r, Hartley, and Stew{\'{e}}nius]{DBLP:conf/cvpr/NisterHS07}
David Nist{\'{e}}r, Richard~I. Hartley, and Henrik Stew{\'{e}}nius.
\newblock Using galois theory to prove structure from motion algorithms are optimal.
\newblock In \emph{2007 {IEEE} Computer Society Conference on Computer Vision and Pattern Recognition {(CVPR} 2007), 18-23 June 2007, Minneapolis, Minnesota, {USA}}. {IEEE} Computer Society, 2007.

\bibitem[Pautrat et~al.(2023{\natexlab{a}})Pautrat, Barath, Larsson, Oswald, and Pollefeys]{DBLP:conf/cvpr/PautratBLOP23}
R{\'{e}}mi Pautrat, Daniel Barath, Viktor Larsson, Martin~R. Oswald, and Marc Pollefeys.
\newblock Deeplsd: Line segment detection and refinement with deep image gradients.
\newblock In \emph{{IEEE/CVF} Conference on Computer Vision and Pattern Recognition, {CVPR} 2023, Vancouver, BC, Canada, June 17-24, 2023}, pages 17327--17336. {IEEE}, 2023{\natexlab{a}}.

\bibitem[Pautrat et~al.(2023{\natexlab{b}})Pautrat, Su{\'a}rez, Yu, Pollefeys, and Larsson]{pautrat2023gluestick}
R{\'e}mi Pautrat, Iago Su{\'a}rez, Yifan Yu, Marc Pollefeys, and Viktor Larsson.
\newblock Gluestick: Robust image matching by sticking points and lines together.
\newblock In \emph{Proceedings of the IEEE/CVF International Conference on Computer Vision}, pages 9706--9716, 2023{\natexlab{b}}.

\bibitem[Persson and Nordberg(2018)]{DBLP:conf/eccv/PerssonN18}
Mikael Persson and Klas Nordberg.
\newblock Lambda twist: An accurate fast robust perspective three point {(P3P)} solver.
\newblock In \emph{Computer Vision - {ECCV} 2018 - 15th European Conference, Munich, Germany, September 8-14, 2018, Proceedings, Part {IV}}, pages 334--349. Springer, 2018.

\bibitem[Raguram et~al.(2013)Raguram, Chum, Pollefeys, Matas, and Frahm]{DBLP:journals/pami/RaguramCPMF13}
Rahul Raguram, Ondrej Chum, Marc Pollefeys, Jiri Matas, and Jan{-}Michael Frahm.
\newblock {USAC:} {A} universal framework for random sample consensus.
\newblock \emph{{IEEE} Trans. Pattern Anal. Mach. Intell.}, 35\penalty0 (8):\penalty0 2022--2038, 2013.

\bibitem[Ramalingam et~al.(2011)Ramalingam, Bouaziz, and Sturm]{DBLP:conf/icra/RamalingamBS11}
Srikumar Ramalingam, Sofien Bouaziz, and Peter~F. Sturm.
\newblock Pose estimation using both points and lines for geo-localization.
\newblock In \emph{{IEEE} International Conference on Robotics and Automation, {ICRA} 2011, Shanghai, China, 9-13 May 2011}, pages 4716--4723. {IEEE}, 2011.

\bibitem[Sattler et~al.(2017)Sattler, Leibe, and Kobbelt]{DBLP:journals/pami/SattlerLK17}
Torsten Sattler, Bastian Leibe, and Leif Kobbelt.
\newblock Efficient {\&} effective prioritized matching for large-scale image-based localization.
\newblock \emph{{IEEE} Trans. Pattern Anal. Mach. Intell.}, 39\penalty0 (9):\penalty0 1744--1756, 2017.

\bibitem[Sch{\"{o}}nberger and Frahm(2016)]{DBLP:conf/cvpr/SchonbergerF16}
Johannes~L. Sch{\"{o}}nberger and Jan{-}Michael Frahm.
\newblock Structure-from-motion revisited.
\newblock In \emph{2016 {IEEE} Conference on Computer Vision and Pattern Recognition, {CVPR} 2016, Las Vegas, NV, USA, June 27-30, 2016}, pages 4104--4113. {IEEE} Computer Society, 2016.

\bibitem[Ventura et~al.(2014)Ventura, Arth, Reitmayr, and Schmalstieg]{DBLP:journals/tvcg/VenturaARS14}
Jonathan Ventura, Clemens Arth, Gerhard Reitmayr, and Dieter Schmalstieg.
\newblock Global localization from monocular {SLAM} on a mobile phone.
\newblock \emph{{IEEE} Trans. Vis. Comput. Graph.}, 20\penalty0 (4):\penalty0 531--539, 2014.

\bibitem[Xu et~al.(2017)Xu, Zhang, Cheng, and Koch]{DBLP:journals/pami/XuZCK17}
Chi Xu, Lilian Zhang, Li Cheng, and Reinhard Koch.
\newblock Pose estimation from line correspondences: {A} complete analysis and a series of solutions.
\newblock \emph{{IEEE} Trans. Pattern Anal. Mach. Intell.}, 39\penalty0 (6):\penalty0 1209--1222, 2017.

\bibitem[Zhou et~al.(2018)Zhou, Ye, and Kaess]{DBLP:conf/accv/ZhouYK18}
Lipu Zhou, Jiamin Ye, and Michael Kaess.
\newblock A stable algebraic camera pose estimation for minimal configurations of 2d/3d point and line correspondences.
\newblock In \emph{Computer Vision - {ACCV} 2018 - 14th Asian Conference on Computer Vision, Perth, Australia, December 2-6, 2018, Revised Selected Papers, Part {IV}}, pages 273--288. Springer, 2018.

\end{thebibliography}
}

% WARNING: do not forget to delete the supplementary pages from your submission 
\clearpage
\setcounter{page}{1}
\maketitlesupplementary

\begin{table*}[b!]
    \centering
    \begin{tabular}{r | r r r r | r r }
         %& Model House & Corridor & Merton I & Merton II & Merton III & Library & \\
          & \multicolumn{4}{|c|}{P2P1L} & \multicolumn{2}{c}{P1P2L} \\
         Dataset & OUR & 3Q3 & \textit{Ram.} SVD & \textit{Ram.} LU & OUR & 3Q3 \\
         \midrule
         Model House &0.251,	0.429	&	0.251,	0.429	&	0.251,	0.429	&	0.251,	0.429	&	0.251,	0.429	&	0.251,	0.429 \\
         Corridor & 0.573,	0.580	&	0.573,	0.580	&	0.573,	0.580	&	0.573,	0.580	&	0.573,	0.580	&	0.573,	0.580 \\
         Merton I & 0.005,	3.2e-4	&	0.005,	3.2e-4	&	0.005,	3.2e-4	&	0.005,	3.2e-4	&	0.005,	3.2e-4	&	0.005,	3.2e-4 \\
         Merton II & 0.003,	4.6e-4	&	0.003,	4.6e-4	&	0.003,	4.6e-4	&	0.003,	4.6e-4	&	0.003,	4.6e-4	&	0.003,	4.6e-4 \\
         Merton III & 0.007,	0.001	&	0.007,	0.001	&	0.007,	0.001	&	0.007,	0.001	&	0.007,	0.001	&	0.007,	0.001 \\
         Library & 0.011,	0.002	&	0.011,	0.002	&	0.011,	0.002	&	0.011,	0.002	&	0.011,	0.002	&	0.011,	0.002 \\
         Wadham & 0.007,	0.001	&	0.007,	0.001	&	0.007,	0.001	&	0.007,	0.001	&	0.007,	0.001	&	0.007,	0.001
 \\
    \end{tabular}
    \caption{\textbf{RANSAC error}, on Oxford Multi-view dataset \cite{oxdata_multiview}, in degrees. Every cell shows rotation and translation error. }
    \label{tab:ransac_error}
\end{table*}

\appendix

\section{P2P1L in the coplanar case}\label{sm:p2p1l-coplanar}

Here, we describe the modified version of the P2P1L solver (\cref{subsec:p2p1l}) capable of handling the coplanar case. The input and output to this solver are the same as those of the original P2P1L solver.

The solver is identical to the original P2P1L solver until Equation~\eqref{eq:T-backsolve}. Then, we express values $R_{11}, R_{31}, R_{22}$ in terms of $R_{21}, R_{23}$ as
\begin{equation} \label{eq:linear_modified}
    \normalsize
    \begin{split}
        R_{11} = \frac{b_2 a_1 - b_1 a_2}{ b_1 b_2 X_2 (Y_3-Y_4) }  \\
        \cdot \Bigg( \bigg(Y_3 X_4 - Y_4 X_3 + \frac{b_1 a_2 (X_2 Y_3 - X_1)}{b_2 a_1 - b_1 a_2} \bigg) R_{21} \\ + Z_4 Y_3 R_{23}   \Bigg)
        \\[0.5em]
        R_{31} = \Bigg( \frac{b_2-b_1}{b_1 b_2 X_2 (Y_3-Y_4)} \bigg(Y_3 X_4 - Y_4 X_3 + \vphantom{\int_1^2}  \\  \frac{b_1 a_2 (X_2 Y_3 - X_1)}{b_2 a_1 - b_1 a_2} \bigg)  + \frac{a_1-a_2}{b_2 a_1 - b_1 a_2}  \Bigg) R_{21} \\ + \frac{(b_2-b_1) Z_4 Y_3}{b_1 b_2 X_2 (Y_3-Y_4)} R_{23},
        \\[0.5em]
        R_{22} = \Bigg( \frac{1}{Y_3 (Y_3-Y_4)} \bigg(Y_3 X_4 - Y_4 X_3 \vphantom{\int_1^2} + \\ \frac{b_1 a_2 (X_2 Y_3 - X_1)}{b_2 a_1 - b_1 a_2} \bigg)  - \frac{X_3}{Y_3} - \frac{b_1 a_2 X_2}{Y_3 (b_2 a_1 - b_1 a_2)} \Bigg) R_{21} \\ + \frac{Z_4}{Y_3-Y_4} R_{23}
    \end{split}
    \normalsize
\end{equation}

Then, we substitute \eqref{eq:linear_modified} into~\eqref{eq:ortho},
to obtain two bivariate quadratic constraints in $R_{2 1}$ and $R_{2 3}$. We can write them in matrix form as
\begin{equation}\label{eq:two-quadratics-modified}
\begin{pmatrix}
c_1 & c_2 & c_3 \\
d_1 & d_2 & d_3
\end{pmatrix}
\cdot 
\begin{pmatrix}
R_{2 1}^2 \\
R_{2 1} R_{2 3} \\
R_{2 3}^2 
\end{pmatrix}
= 
\begin{pmatrix}
1 \\
1
\end{pmatrix}
,
\end{equation}
where the coefficients $c_{1},c_2,c_3,d_1,d_2,d_3$ are rational functions of the problem data. Similarly to the original solver, we apply the change of variables 
\begin{equation}\label{eq:cov_modified}
u = R_{2 1}^2, \quad v = R_{2 3} / R_{2 1}
\end{equation}

Subtracting the two equations in~\eqref{eq:two-quadratics-modified}, we obtain
\[
(c_1 - d_1) u + (c_2-d_2) uv + (c_3 - d_3) u v^2 = 0.
\]
Assuming $u\ne 0,$ we therefore have the univariate quadratic equation in $v$
\begin{equation}\label{eq:v-quadratic_modified}
(c_1 - d_1) + (c_2-d_2) v + (c_3 - d_3) v^2 = 0.
\end{equation}

We recover value $v$ as one of the roots of \eqref{eq:v-quadratic_modified} and $u$ as
\begin{equation}\label{eq:u-linear}
u = (c_1 + c_2 v + c_3 v^2)^{-1},
\quad 
\text{or}
\quad 
u = (d_1 + d_2 v + d_3 v^2)^{-1}.
\end{equation}

After the values $u$, $v$ are recovered, we obtain $R_{21}, R_{23}$ according to \eqref{eq:two-quadratics-modified} and $R_{11}, R_{31}, R_{22}$ according to \eqref{eq:linear_modified}. The rest of the solver is identical to the original solver.

\section{Evaluation of the pose error. }

In this section, we show the pose estimation errors obtained in the RANSAC experiment, following the same experimental setup as in \cref{subsec:ransac_experiments}.
The results are presented in Table~\ref{tab:ransac_error}.
%, show that  shows the final pose estimation errors in RANSAC. 
As expected, the errors are the same for all solvers. This demonstrates that our solvers can achieve improved runtimes without sacrificing any accuracy.

\clearpage

%\section{Rationale}
%\label{sec:rationale}
% 
%Having the supplementary compiled together with the main paper means that:
% 
%\begin{itemize}
%\item The supplementary can back-reference sections of the main paper, for example, we can refer to \cref{sec:intro};
%\item The main paper can forward reference sub-sections within the supplementary explicitly (e.g. referring to a particular experiment); 
%\item When submitted to arXiv, the supplementary will already included at the end of the paper.
%\end{itemize}
% 
%To split the supplementary pages from the main paper, you can use \href{https://support.apple.com/en-ca/guide/preview/prvw11793/mac#:~:text=Delete%20a%20page%20from%20a,or%20choose%20Edit%20%3E%20Delete).}{Preview (on macOS)}, \href{https://www.adobe.com/acrobat/how-to/delete-pages-from-pdf.html#:~:text=Choose%20%E2%80%9CTools%E2%80%9D%20%3E%20%E2%80%9COrganize,or%20pages%20from%20the%20file.}{Adobe Acrobat} (on all OSs), as well as \href{https://superuser.com/questions/517986/is-it-possible-to-delete-some-pages-of-a-pdf-document}{command line tools}.

\end{document}